%% file: main.tex
\documentclass[%
  letterpaper,%
  twocolumn,%
  english%
]{article}

\usepackage[%
  textwidth=492pt,%
  textheight=624pt,%
  columnsep=12pt,%
  footskip=48pt,%
  footnotesep=36pt%
]{geometry}
\usepackage[big,raggedright,compact]{titlesec}
\usepackage{fancyhdr}
\usepackage{amssymb}
\usepackage{amsthm}
\usepackage[p]{newtx}
\usepackage{etoolbox}
\usepackage[super,comma,sort]{natbib}

\usepackage[hidelinks]{hyperref}
\usepackage{caption}
\usepackage{subcaption}
\usepackage{booktabs}
\usepackage{pgf}
\usepackage{pgfplots}
\usepackage{tikz}
\usepackage{tikzscale}
\usepackage{svg}

\providecommand{\tightlist}{%
  \setlength{\itemsep}{0pt}\setlength{\parskip}{0pt}%
}

\let\oldfootnote\footnote
\renewcommand{\footnote}[1]{\ifx\caption@checkhyp\caption@refhypmode\else\protect\oldfootnote{#1}\fi}

\usepackage[Export]{adjustbox}
\usepackage{environ}
\usepackage{framed}


\newcommand{\maxfigwidth}{\textwidth}
\newcommand{\maxfigheight}{\textheight}
\newcommand{\maxtabwidth}{\textwidth}
\newcommand{\maxtabheight}{\textheight}

\let\origincgfx\includegraphics
\renewcommand{\includegraphics}[2][]{%
\origincgfx[%
  #1,keepaspectratio,max width=\maxfigwidth,max height=\maxfigheight%
]{#2}%
}

\def\mdccap{CAPTION}
\def\mdclab{LABEL}

\RenewEnviron{figure}{
  \let\oldcaption\caption
  \let\oldsetcaptionsubtype\setcaptionsubtype
  \let\oldlabel\label

  \renewcommand{\setcaptionsubtype}{}
  \renewcommand{\caption}[2][]{\global\def\mdccap{##2}}
  \renewcommand{\label}[1]{%
    \ifdefined\insubfloat
      \oldlabel{##1}
    \else
      \global\def\mdclab{##1}
    \fi
  }

  \let\oldsubfloat\subfloat
  \renewcommand{\subfloat}[2][]{\def\insubfloat{}\oldsubfloat[##1]{##2}\let\insubfloat\undefined}

  \sbox{0}{\BODY}

  \let\setcaptionsubtype\oldsetcaptionsubtype

  \xdef\thelinew{\the\linewidth}

  \ifdim\wd0>\thelinew
    \begin{figure*}
  \else
    \begin{oldfigure}
  \fi

      \setlength{\FrameRule}{0pt}
      \setlength{\FrameSep}{0pt}

      \begin{framed}
      \begin{adjustbox}{center}
      \begin{adjustbox}{max totalheight=\maxfigheight}
        \begin{minipage}{\maxfigwidth}
          \centering
          \BODY
        \end{minipage}
      \end{adjustbox}
      \end{adjustbox}
      \end{framed}

      \oldcaption{\mdccap}
      \oldlabel{\mdclab}

  \ifdim\wd0>\thelinew
    \end{figure*}
  \else
    \end{oldfigure}
  \fi

  \let\caption\oldcaption
  \let\label\oldlabel
}

\RenewEnviron{table}{
  \let\oldcaption\caption
  \renewcommand{\caption}[2][]{\global\def\mdccap{##2}}

  \sbox{0}{\BODY}

  \xdef\thelinew{\the\linewidth}

  \ifdim\wd0>\thelinew
    \begin{table*}
  \else
    \begin{oldtable}
  \fi

      \begin{adjustbox}{center}
        \maxsizebox{\maxtabwidth}{\maxtabheight}{\BODY}
      \end{adjustbox}

      \oldcaption{\mdccap}

  \ifdim\wd0>\thelinew
    \end{table*}
  \else
    \end{oldtable}
  \fi
}

\usepackage{stfloats}


\usepackage{mathtools}
\usepackage{mleftright}
\usepackage{bm}
\usepackage{array}

\renewcommand\left\mleft
\renewcommand\right\mright

\makeatletter
\def\resetMathstrut@{%
  \setbox\z@\hbox{%
    \mathchardef\@tempa\mathcode`\[\relax
    \mathchardef\@tempc\mathcode`\]\relax
    \def\@tempb##1"##2##3{\the\textfont"##3\char"}%
    \expandafter\@tempb\meaning\@tempa \relax
  }%
  \ht\Mathstrutbox@\ht\z@ \dp\Mathstrutbox@\dp\z@}
\makeatother
\begingroup
  \catcode`(\active \xdef({\left\string(}
  \catcode`)\active \xdef){\right\string)}
\endgroup
\mathcode`(="8000 \mathcode`)="8000

\usepackage{xparse}
\usepackage{letltxmacro}

\usepackage[T1]{fontenc}
\usepackage[babel=true,tracking=true]{microtype}
\usepackage{babel}
\usepackage[title]{appendix}
\usepackage{footnote}
\usepackage[hang,bottom,ragged]{footmisc}
\usepackage{acronym}
\usepackage{relsize}
\usepackage{xcolor}
\usepackage{fancyvrb}
\usepackage[defaultlines=2,all]{nowidow}

\makesavenoteenv{table}
\makesavenoteenv{table*}
\makesavenoteenv{figure}
\makesavenoteenv{figure*}

\makeatletter
\renewcommand\@makefnmark{\mbox{\textsuperscript{\@thefnmark}\,}}
\makeatother

\makeatletter
\def\blfootnote{\gdef\@thefnmark{}\@footnotetext}
\makeatother

\hypersetup{
  pdfinfo={
    Title={AlphaNet: Improving Long-Tail Classification By Combining Classifiers},
    Author={Nadine Chang, Jayanth Koushik, Aarti Singh, Martial Hebert, Yu-Xiong Wang, Michael J. Tarr}
  }
}

\AtBeginEnvironment{tabular}{\smaller}

\renewcommand*{\thefootnote}{\alph{footnote}}

\titlelabel{\thetitle.\enskip}

\makeatletter
\def\@maketitle{%
  \newpage
  \begin{flushleft}%
    \normalfont%
    {\LARGE \@title \par}%
    \vskip 1.5em%
    {\large
      \lineskip .5em%
      \begin{tabular}[t]{@{}l@{}}%
          \@author
      \end{tabular}\par}%
    {\normalsize \@date}%
  \end{flushleft}%
  \par
  \vskip 1.5em}
\makeatother

\makeatletter
\def\and{%
  \end{tabular}%
  \hskip 1em \@plus.17fil%
  \begin{tabular}[t]{@{}l@{}}}%
\makeatother

\title{AlphaNet: Improving Long-Tail Classification By Combining Classifiers}
\date{}

\author{%
      \mbox{Nadine Chang\textsuperscript{\(\bm{\dag}\),*}}\\%
    \mbox{\href{mailto:nchang1@cs.cmu.edu}{\texttt{\textsmaller{nchang1@cs.cmu.edu}}}}
  \and
      \mbox{Jayanth Koushik\textsuperscript{\(\bm{\dag}\),*}}\\%
    \mbox{\href{mailto:jkoushik@andrew.cmu.edu}{\texttt{\textsmaller{jkoushik@andrew.cmu.edu}}}}
  \and
      \mbox{Aarti Singh\textsuperscript{\(\bm{\dag}\)}}\\%
    \mbox{\href{mailto:aartisingh@cmu.edu}{\texttt{\textsmaller{aartisingh@cmu.edu}}}}
  \and
      \mbox{Martial Hebert\textsuperscript{\(\bm{\dag}\)}}\\%
    \mbox{\href{mailto:hebert@cs.cmu.edu}{\texttt{\textsmaller{hebert@cs.cmu.edu}}}}
  \and
      \mbox{Yu-Xiong Wang\textsuperscript{\(\bm{\ddag}\)}}\\%
    \mbox{\href{mailto:yxw@illinois.edu}{\texttt{\textsmaller{yxw@illinois.edu}}}}
  \and
      \mbox{Michael J. Tarr\textsuperscript{\(\bm{\dag}\)}}\\%
    \mbox{\href{mailto:michaeltarr@cmu.edu}{\texttt{\textsmaller{michaeltarr@cmu.edu}}}}
}

\input{src/include_commands}

\input{src/include_debugcmd}

\begin{document}

\maketitle

\thispagestyle{fancy}
\parindent=2em

\blfootnote{%
  {%
    \normalfont\scriptsize%
    This document can be read online at \href{https://jkoushik.me/alphanet}{\texttt{https://jkoushik.me/alphanet}}.\\%
    \mbox{\textsuperscript{*}\,Equal contribution.}\enskip%
    \mbox{\textsuperscript{\(\bm{\dag}\)}\,Carnegie Mellon University.}\enskip\mbox{\textsuperscript{\(\bm{\ddag}\)}\,University of Illinois Urbana-Champaign.}\enskip%
  }%
}

\hrule height 1pt
\vspace{1ex}%
{\parindent=0pt \normalfont\normalsize \input{src/abstract}}
\vspace{1ex}%
\hrule height 1pt
\vspace{1ex}

\input{src/body}

\input{src/section_1_intro}
\input{src/section_2_related}
\input{src/section_3_method}
\input{src/section_4_experiments}
\input{src/section_5_conclusion}
\input{src/section_6_acknowledgements}

\footnotesize
\bibliographystyle{naturemag}
\bibliography{references.bib}
\normalsize

\clearpage
\begin{appendices}
  \renewcommand{\thefigure}{A\arabic{figure}}
  \renewcommand{\thetable}{A\arabic{table}}
  \renewcommand{\thefootnote}{A\arabic{footnote}}
  \setcounter{table}{0}
  \setcounter{figure}{0}
  \setcounter{footnote}{0}
  \input{src/appendix_implementation}
\end{appendices}

\clearpage
\begin{appendices}
  \renewcommand{\thefigure}{B\arabic{figure}}
  \renewcommand{\thetable}{B\arabic{table}}
  \renewcommand{\thefootnote}{B\arabic{footnote}}
  \setcounter{table}{0}
  \setcounter{figure}{0}
  \setcounter{footnote}{0}
  \input{src/appendix_ksweep}
\end{appendices}

\clearpage
\begin{appendices}
  \renewcommand{\thefigure}{C\arabic{figure}}
  \renewcommand{\thetable}{C\arabic{table}}
  \renewcommand{\thefootnote}{C\arabic{footnote}}
  \setcounter{table}{0}
  \setcounter{figure}{0}
  \setcounter{footnote}{0}
  \input{src/appendix_rhosweep}
\end{appendices}

\clearpage
\begin{appendices}
  \renewcommand{\thefigure}{D\arabic{figure}}
  \renewcommand{\thetable}{D\arabic{table}}
  \renewcommand{\thefootnote}{D\arabic{footnote}}
  \setcounter{table}{0}
  \setcounter{figure}{0}
  \setcounter{footnote}{0}
  \input{src/appendix_perclsdels}
\end{appendices}

\end{document}

%% file: src/include_commands.tex
\newcommand{\R}{\mathbb{R}}

\newcommand{\suml}{\sum\limits}
\newcommand{\sumnl}{\sum\nolimits}

\newcommand{\abs}[1]{\left|{#1}\right|}
\newcommand{\norm}[1]{\left\|{#1}\right\|}

\newcommand{\set}[1]{\left\{#1\right\}}

\renewcommand{\c}[1]{\overline{#1}}
\renewcommand{\v}[1]{\bm{#1}}
\newcommand{\shat}[1]{\vphantom{#1}\smash[t]{\hat{#1}}}

%% file: src/include_debugcmd.tex
\newcommand{\pernote}[2]{}

%% file: src/abstract.tex
Methods in long-tail learning focus on improving performance for data-poor (rare) classes; however, performance for such classes remains much lower than performance for more data-rich (frequent) classes. Analyzing the predictions of long-tail methods for rare classes reveals that a large number of errors are due to misclassification of rare items as visually similar frequent classes. To address this problem, we introduce AlphaNet, a method that can be applied to existing models, performing post hoc correction on classifiers of rare classes. Starting with a pre-trained model, we find frequent classes that are closest to rare classes in the model's representation space and learn weights to update rare class classifiers with a linear combination of frequent class classifiers. AlphaNet, applied to several models, greatly improves test accuracy for rare classes in multiple long-tailed datasets, with very little change to overall accuracy. Our method also provides a way to control the trade-off between rare class and overall accuracy, making it practical for long-tail classification in the wild.

%% file: src/body.tex
\graphicspath{{figures/}}

\raggedbottom

%% file: src/section_1_intro.tex
\hypertarget{sec:intro}{%
\section{Introduction}\label{sec:intro}}

The significance of long-tailed distributions in real-world applications (such as autonomous driving\citep{2022.Anguelov.Jiang}, and medical image analysis\citep{2022.Bian.Yang}) has spurred a variety of approaches for long-tail classification\citep{2022.Guo.Yang}. Learning in this setting is challenging because many classes are ``rare'' -- having only a small number of training samples. Some methods re-sample more data for rare classes in an effort to address data imbalances\citep{2019.Belongie.Cui, 2019.Ma.Cao}, while other methods adjust learned classifiers to re-weight them in favor of rare classes\citep{2020.Kalantidis.Kang}. Both re-sampling and re-weighting methods provide strong baselines for long-tail classification tasks. However, state-of-the-art results are achieved by more complex methods that, for example, learn multiple experts\citep{2021.Yu.Wang, 2021.Hwang.Cai}, perform multi-stage distillation\citep{2021.Wu.Li}, or use a combination of weight decay, loss balancing, and norm thresholding\citep{2022.Kong.Alshammari}.

Despite these advances, accuracy on rare classes continues to be significantly lower than overall accuracy. For example, on ImageNet‑LT -- a long-tailed dataset sampled from ImageNet\citep{2009.Fei-Fei.Deng} -- the 6-expert ensemble RIDE model\citep{2021.Yu.Wang} has an average accuracy of 68.9\% on frequent classes, but an average accuracy of 36.5\% on rare classes.\footnote{Results for the 6-expert model are presented in the GitHub repository for the original paper at \href{https://github.com/frank-xwang/RIDE-LongTailRecognition/blob/main/MODEL_ZOO.md}{\texttt{github.com/frank-xwang/\ RIDE-LongTailRecognition/blob/main/MODEL\_ZOO.md}}.} In addition to reducing overall accuracy, such performance imbalances raise ethical concerns in contexts where unequal accuracy leads to biased outcomes, such as medical imaging\citep{2022.Ferrante.Lara}, or face detection\citep{2018.Gebru.Buolamwini}. For instance, models trained on chest X‑ray images consistently under-diagnosed minority groups\citep{2021.Ghassemi.Seyyed}, and similarly, cardiac image segmentation showed significant differences between racial groups\citep{2021.King.Puyol}.

\begin{figure}
\subfloat[Predictions from cRT model on test samples from `few' split of ImageNet‑LT. For a misclassified sample, if the predicted class is one of the 5 `base' split nearest neighbors (NNs) of the true class, it is considered to be incorrectly classified as a NN. A large number of samples are misclassified in this way.]{\includegraphics{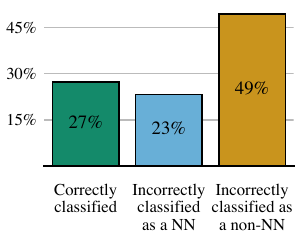}\label{fig:analysis:bins}}
\quad
\subfloat[Sample images from two classes in ImageNet‑LT. `Lhasa' is a `few' split class, and `Tibetan terrier' is a `base' split class. The classes are visually very similar, leading to misclassifications.]{\includegraphics{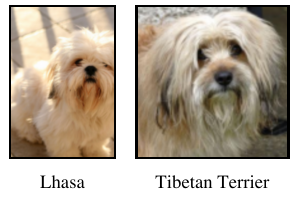}\label{fig:analysis:egs}}
\quad
\subfloat[Per-class test accuracy of cRT model on `few' split of ImageNet‑LT, versus the mean Euclidean distance to 5 nearest neighbor (NN) `base' split classes. The line is a bootstrapped linear regression fit, and `\(r\)' (top right) is Pearson correlation. There is a high correlation, i.e., `few' split classes with close `base' split NNs are more likely to be misclassified.]{\includegraphics{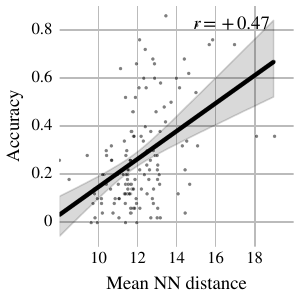}\label{fig:analysis:acc_vs_dist}}
\caption{Analysis of `few' split predictions on ImageNet‑LT.}
\label{fig:analysis}
\end{figure}

To understand the poor rare class performance of long-tail models, we analyzed predictions of the cRT model\citep{2020.Kalantidis.Kang} on test samples from ImageNet‑LT's `few' split (i.e., classes with limited training samples). Figure~\ref{fig:analysis:bins} shows predictions binned into three groups: (1) samples classified correctly; (2) samples incorrectly classified as a visually similar `base' split\footnote{The `base' split is the complement of the `few' split, composed of classes with many training samples.} class (e.g., `husky' instead of `malamute'); and (3) samples incorrectly classified as a visually dissimilar class (e.g., `goldfish' instead of `malamute'). A significant portion of the misclassifications (about 23\%) are to visually similar frequent classes. Figure~\ref{fig:analysis:egs} highlights the reason behind this issue, with samples from one pair of visually similar classes; the differences are subtle, and can be hard even for humans to identify. To get a quantitative understanding, we analyzed the relationship between per-class test accuracy and mean distance of a class to its nearest neighbors (see Section~\ref{sec:method} for details). Figure~\ref{fig:analysis:acc_vs_dist} shows a strong positive correlation between accuracy and mean distance, meaning that rare classes with close neighbors have lower test accuracy than classes with distant neighbors.

Based on these analyses, we designed a method to directly improve the accuracy on rare classes in long-tail classification. Our method, AlphaNet, uses information from visually similar frequent classes to improve classifiers for rare classes. Figure~\ref{fig:alphanet} illustrates the pipeline of our method. At a high level, AlphaNet can be seen as moving the classifiers for rare classes based on their position relative to visually similar classes. Importantly, AlphaNet updates classifiers without making any changes to the representation space, or to other classifiers in the model. It performs a post hoc correction, and as such, is applicable to use cases where existing base classifiers are either unavailable or fixed (e.g., due to commercial interests or data privacy protections). The simplicity of our method lends to computational advantages -- AlphaNet can be trained rapidly, and on top of any classification model. We will demonstrate that AlphaNet, applied to a variety of long-tail classification models, significantly improves rare class accuracy on multiple datasets.

%% file: src/section_2_related.tex
\hypertarget{sec:relwork}{%
\section{Related work}\label{sec:relwork}}

Our work falls in the domain of long-tail learning, where the distribution of class sizes -- measured via number of training samples -- models that of the visual world; many classes have only a few samples, while a small number have many\citep{2023.Feng.Zhang}. Kang et~al.\citep{2020.Kalantidis.Kang} established strong baselines on long-tailed datasets by decoupling classifiers and representations. We apply AlphaNet to of their proposed baseline methods: (1) the cRT (classifier re-training) model, which fixes representations, and trains classifiers from scratch; and (2) the LWS (learnable weight scaling) model, which also fixes representations, and only \emph{rescales} classifiers, with scales learned from the training data.

In contrast to the above simple methods, many complex methods have been proposed, and have continued to push the state-of-the-art for long-tail recognition\citep{2019.Yu.Liu, 2021.Yu.Wang, 2021.Wu.Li, 2021.Hwang.Cai, 2022.Kong.Alshammari}. We used two of these methods to evaluate AlphaNet: (1) the RIDE (RoutIng Diverse Experts) model of Wang et~al.\citep{2021.Yu.Wang}, which achieves low bias and variance, by training with a ``distribution-aware diversity loss'' and using multiple experts, respectively; and (2) the LTR (long-tail recognition) model of Alshammari et~al.\citep{2022.Kong.Alshammari}, which uses a combination of class-balanced loss, weight decay, and max-norm regularization -- in this work we will refer to this model simply as the LTR model.

In the rest of this section, we discuss some works that make use of similar ideas as AlphaNet.

\hypertarget{sec:relwork:transfer}{%
\subsection{Knowledge transfer}\label{sec:relwork:transfer}}

AlphaNet bears resemblance to methods that create new classifiers by transferring knowledge from existing classifiers. These methods appear in a number of domains, such as transfer learning, meta-learning, and multi-task learning\citep{2012.Lorien.Thrun, 1997.Caruana, 1997.Wiering.Schmidhuber, 2010.Yang.Pan}. It should be noted, however, that AlphaNet does not create new classifiers -- it only \emph{modifies} existing classifiers by combining them with others.

Pertinent to our problem setting is the work by Wang et~al.\citep{2016.Hebert.Wang}, who showed that a ``generic, category agnostic transformation'' can be learned from models trained on few samples, to models trained on many samples. In our work, we implicitly learn a similar transformation, but with the source and target classifiers within the same model. Additionally, the transformation is constrained to be a linear combination. A similar paradigm was analyzed by Du et~al.\citep{2016.Poczos.Du}, who showed that for cases where the target function is generated by a simple transformation of the source function, there are theoretical performance guarantees for a large class of functions.

\hypertarget{sec:relwork:composition}{%
\subsection{Classifier composition}\label{sec:relwork:composition}}

In low-shot\footnote{Low-shot learning is also referred to as few-shot learning, and as one-shot learning if only a single training example is available per class.} and zero-shot classification, new classifiers are learned using few or zero training examples\citep{2006.Perona.Fei-Fei, 2008.Bengio.Larochelle}. Some methods have done this by directly combining existing classifiers. For example, Mensink et~al.\citep{2014.Snoek.Mensink} learned new classifiers as linear combinations of existing classifiers, with weights determined by co-occurrence statistics. Changpinyo et~al.\citep{2016.Sha.Changpinyo} introduced ``phantom classes'' and used their classifiers as bases to compose new classifiers through convex combination.

The idea of combining existing classifiers was also used by Aytar et~al.\citep{2015.Zisserman.Aytar}, in their work on enhancing single exemplar support vector machines (SVMs). In their method, an extra regularization term is added to the SVM loss function, which encourages the learned classifier to be close to a linear combination of previously learned classifiers. Classifiers trained on image patches are used to transfer knowledge to a classifier trained on a single positive exemplar.

\hypertarget{sec:relwork:composition:boosting}{%
\subsubsection{Boosting}\label{sec:relwork:composition:boosting}}

Composing weak classifiers to build strong classifiers also bears resemblance to the idea of boosting\citep{1990.Schapire}. The popular AdaBoost\citep{1997.Schapire.Freund} algorithm linearly combines classifiers based on single features (e.g., decision stumps), and iteratively re-weights training samples based on their error. For the case of multi-class classification, Torralba et~al.\citep{2007.Freeman.Torralba} built a classifier that combines several binary classifiers, each designed to separate a single class from the others. Their method identifies common features that be shared across classifiers, which reduces the computational load, and the amount of training data required.

It is important to note that boosting methods employ a different form of composition that our methods. Specifically, our focus is on classification methods where the performance on a \emph{subset of classes} is poor. Unlike boosting methods, we do not incorporate additional features -- improvements are made by adjusting classifiers within the learned representation space.

%% file: src/section_3_method.tex
\hypertarget{sec:method}{%
\section{Method}\label{sec:method}}

Our problem setting is multi-class classification with \(C\) classes, where each input has a corresponding class label in \(\set{0, \dots, C - 1}\), and the goal is to learn a mapping from inputs to labels. We are specifically interested in visual recognition; so inputs are images, and classes are object categories. AlphaNet is applied to a pre-trained classification model; we assume that this model can be decoupled into two parts: the first part maps images to feature vectors, and the second part maps feature vectors to ``scores'', one for each of the \(C\) classes. The prediction for an image is the class index with largest corresponding score. Typically (and in all our experiments), for a convolutional network, the feature vector for an image is the output of the penultimate layer, and the last layer is a linear mapping. So, each classifier is a vector, and the score for a class is the dot product of the feature vector with the corresponding classifier. Typically a bias term is present, which is added to the dot product. We do not modify this term, and use it as-is if present.

In this work, we define the distance between two classes as the distance between their average training set representation. Let \(f\) be the function mapping images to feature vectors in a \(d\)-dimensional representation space. For a class \(c\) with \(n^c\) training samples \(I^c_1, \dots, I^c_{n^c}\), let \(\v{z}^c \equiv (1/n^c) \sumnl_i f(I^c_i)\) be the average training set representation. Given a distance function \(\mu: \R^d \times \R^d \to \R\), we define the distance between two classes \(c_1\) and \(c_2\) as \(m_\mu(c_1, c_2) \equiv \mu(\v{z}^{c_1}, \v{z}^{c_2})\).

Given a long-tailed dataset, the `few' split (\(C^F\)), is defined as the set of classes with fewer than \(T\) training samples, for some constant \(T\) (equal to 20 for the datasets used in this work). The remaining classes form the `base' split (\(C^B\)). AlphaNet is used to update the `few' split classifiers using nearest neighbors from the `base' split.

\hypertarget{sec:method:impl}{%
\subsection{AlphaNet implementation}\label{sec:method:impl}}

\begin{figure}
\hypertarget{fig:alphanet}{%
\centering
\includegraphics[width=7.25in,height=\textheight]{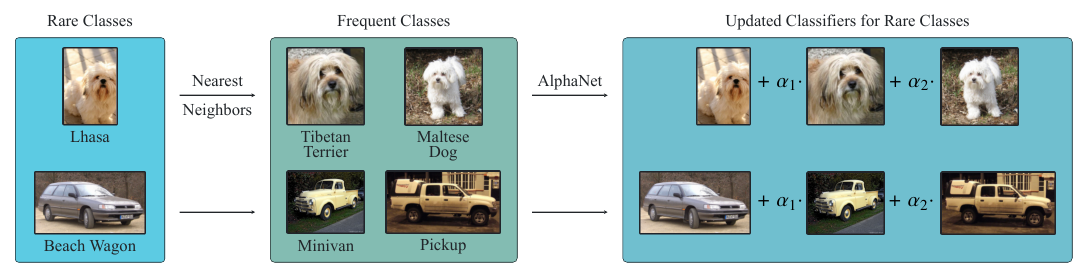}
\caption{Pipeline for AlphaNet. Given a rare class, we identify the nearest neighbor frequent classes based on visual similarity, and then update the rare class' classifier using learned coefficients. One coefficient, \(\alpha\), is learned for each nearest neighbor. The result is an improved classifier for the rare class.}\label{fig:alphanet}
}
\end{figure}

Figure~\ref{fig:alphanet} shows the pipeline of our method. Given a `few' split class \(c\) with classifier \(\v{w}^c\), we find its \(k\) nearest `base' split neighbors based on \(m_\mu\). Let these neighbors have classifiers \(\v{v}^c_1, \dots, \v{v}^c_k\), which are concatenated together into a vector \(\c{\v{v}}^c\). AlphaNet maps \((\v{w}^c, \c{\v{v}}^c)\) to a set of coefficients \(\alpha^c_1, \dots, \alpha^c_k\). The \(\alpha\) coefficients (denoted together as a vector \(\v{\alpha}^c\)), are then scaled to unit 1-norm to obtain \(\tilde{\v{\alpha}}^c\) (the justification for this will be presented shortly): \begin{equation}\protect\hypertarget{eq:alpha_scaling}{}{
       \tilde{\v{\alpha}}^c
\equiv \v{\alpha}^c / \norm{\v{\alpha}^c}_1.
}\label{eq:alpha_scaling}\end{equation} The scaled coefficients are used to update the `few' split classifier (\(\v{w}^c \to \shat{\v{w}}^c\)) through a linear combination: \begin{equation}\protect\hypertarget{eq:alphanet_update}{}{
       \shat{\v{w}}^c
\equiv \v{w}^c + \suml_{i=1}^k \tilde{\alpha}^c_i \v{v}^c_i.
}\label{eq:alphanet_update}\end{equation} Due to the 1-norm scaling, we have \begin{equation}\protect\hypertarget{eq:wdelta_norm_bound}{}{
\begin{aligned}
       \norm{\shat{\v{w}}^c - \v{w}^c}_2
&\le   \suml_{i=1}^k \abs{\tilde{\alpha}^c_i} \cdot \norm{\v{v}^c_i}_2
 \quad \text{(Cauchy-Schwarz inequality)} \\
&\le \max_{i=1,\dots,k} \norm{\v{v}^c_i}_2 \suml_{i=1}^k \abs{\tilde{\alpha}^c_i} \\
&=   \max_{i=1,\dots,k} \norm{\v{v}^c_i}_2 \norm{\tilde{\v{\alpha}}^c}_1 \\
&=   \max_{i=1,\dots,k} \norm{v^c_i}_2,
\end{aligned}
}\label{eq:wdelta_norm_bound}\end{equation} that is, a classifier's change is bound by its nearest neighbors. Thanks to this, we do not need to rescale `base' split classifiers, which may not be possible in certain domains.

Note that a single network is used to generate coefficients for every `few' split class. So, once trained, AlphaNet can be applied even to classes not seen during training. This will be explored in future work.

\hypertarget{sec:method:training}{%
\subsection{Training}\label{sec:method:training}}

The trainable component of AlphaNet is a network with parameters \(\v{\theta}\), which maps \((\v{w}^c, \c{\v{v}}^c)\) to \(\v{\alpha}^c\). We use the original classifier biases, \(\v{b}\) (one per class). So, given a training image \(I\), the per-class prediction scores are given by \begin{equation}\protect\hypertarget{eq:pred_scores}{}{
s(c; I) = \begin{cases}
       f(I)^T \shat{\v{w}}^c + b_c & c \in C^F. \\
       f(I)^T \v{w}^c + b_c        & c \in C^B.
\end{cases}
}\label{eq:pred_scores}\end{equation} That is, class scores are unchanged for `base' split classes, and are computed using updated classifiers for `few' split classes. These scores are used to compute the sample loss (softmax cross-entropy in our experiments), a differentiable function of \(\v{\theta}\). So, \(\v{\theta}\) can be learned using a gradient based optimizer, from mini-batches of training samples.

%% file: src/section_4_experiments.tex
\hypertarget{sec:exp}{%
\section{Experiments}\label{sec:exp}}

\hypertarget{sec:exp:setup}{%
\subsection{Experimental setup}\label{sec:exp:setup}}

A detailed description of the experimental methods is contained in Section~\ref{sec:impl}. A short summary is presented here.

\textbf{Datasets.} We evaluated AlphaNet using three long-tailed datasets:\footnote{Another popular dataset for evaluating long-tail models is iNaturalist\citep{2018.Belongie.Horn}. However, models are able to achieve much more balanced results on this dataset, compared to other long-tailed datasets. For example, with the cRT model, `few' split accuracy (69.2\%) is only 2 points lower than the overall accuracy (71.2\%). So the dataset does not represent a valid use case for our proposed method, and we omitted the dataset from our main experiments. Results for this dataset are included in the appendix (Section~\ref{sec:rhosweep}).} ImageNet‑LT and Places‑LT, curated by Liu et~al.\citep{2019.Yu.Liu} and CIFAR‑100‑LT, created using the procedure described by Cui et~al.\citep{2019.Belongie.Cui}. These datasets are sampled from their respective original datasets -- ImageNet\citep{2015.Fei-Fei.Russakovsky}, Places365\citep{2018.Torralba.Zhou}, and CIFAR‑100\citep{2009.Krizhevsky} -- such that the number of per-class training samples has a long-tailed distribution.

The datasets are broken down into three broad splits based on the number of training samples per class: (1) `many' contains classes with greater than 100 samples; (2) `medium' contains classes with greater than or equal to 20 samples but less than or equal to 100 samples; and (3) `few' contains classes with fewer than 20 samples. \emph{The test set is always balanced}, containing an equal number of samples for each class. We refer to the combined `many' and `medium' splits as the `base' split.

\textbf{Training data sampling.} In order to prevent over-fitting on the `few' split samples, we used a class balanced sampling approach, using all `few' split samples, and a portion of the `base' split samples. Given \(F\) `few' split samples and a ratio \(\rho\), \(\rho F\) samples were drawn from the `base' split every epoch, with sample weights inversely proportional to the size of their class. This ensured that all `base' classes had an equal probability of being sampled.\footnote{For example, suppose there are 2 `base' classes -- class~1 has 10 samples, and class~2 has 100 samples. Then, each class~1 sample is assigned a weight of 0.1, and each class~2 sample is assigned a weight of 0.01. Sampling with this weight distribution, both classes have a 50\% chance of being sampled.} As we show in the following section, \(\rho\) allows us to control the balance between `few' and `base' split accuracy.

\textbf{Training.} All experiments used an AlphaNet with three 32 unit layers. Unless stated otherwise, Euclidean distance was used to find \(k=5\) nearest neighbors for each `few' split class. In this section, we show results for \(\rho\) in \(\set{0.5, 1, 1.5}\). Results for a larger set of \(\rho\)s are shown in Section~\ref{sec:rhosweep}. All experiments were repeated 10 times, and we report average results.

\hypertarget{sec:exp:classres}{%
\subsection{Long-tail classification results}\label{sec:exp:classres}}

\hypertarget{tbl:datasets_baselines_split_accs_vs_rho}{}
\begin{table}
\centering
\begin{tabular}[]{@{}lrrrr@{}}

\toprule\noalign{}
Method & Few & Med. & Many & Overall \\
\midrule\noalign{}

\textbf{ImageNet‑LT} & & & & \\
NCM & \(28.1\) & \(45.3\) & \(56.6\) & \(47.3\) \\
\(\tau\)‑normalized & \(30.7\) & \(46.9\) & \(59.1\) & \(49.4\) \\
& & & & \\
cRT & \(27.4\) & \(46.2\) & \(61.8\) & \(49.6\) \\
\(\alpha\)‑cRT & & & & \\
\(\rho=0.5\) & \(39.7^{1.42}\) & \(42.0^{0.66}\) & \(58.3^{0.52}\) & \(48.0^{0.37}\) \\
\(\rho=1\) & \(34.6^{1.88}\) & \(43.7^{0.51}\) & \(59.7^{0.43}\) & \(48.6^{0.24}\) \\
\(\rho=1.5\) & \(32.6^{2.46}\) & \(44.4^{0.49}\) & \(60.3^{0.38}\) & \(48.9^{0.19}\) \\
& & & & \\
LWS & \(30.4\) & \(47.2\) & \(60.2\) & \(49.9\) \\
\(\alpha\)‑LWS & & & & \\
\(\rho=0.5\) & \(46.9^{0.98}\) & \(38.6^{0.87}\) & \(52.9^{0.86}\) & \(45.3^{0.69}\) \\
\(\rho=1\) & \(41.6^{1.61}\) & \(42.2^{0.53}\) & \(56.0^{0.32}\) & \(47.4^{0.30}\) \\
\(\rho=1.5\) & \(40.1^{1.99}\) & \(43.2^{0.98}\) & \(56.9^{0.76}\) & \(48.0^{0.53}\) \\
& & & & \\
& & & & \\
\textbf{Places‑LT} & & & & \\
NCM & \(27.3\) & \(37.1\) & \(40.4\) & \(36.4\) \\
\(\tau\)‑normalized & \(31.8\) & \(40.7\) & \(37.8\) & \(37.9\) \\
& & & & \\
cRT & \(24.9\) & \(37.6\) & \(42.0\) & \(36.7\) \\
\(\alpha\)‑cRT & & & & \\
\(\rho=0.5\) & \(31.0^{0.88}\) & \(34.5^{0.17}\) & \(40.4^{0.29}\) & \(35.9^{0.09}\) \\
\(\rho=1\) & \(27.0^{1.02}\) & \(36.1^{0.31}\) & \(41.3^{0.13}\) & \(36.2^{0.10}\) \\
\(\rho=1.5\) & \(25.5^{0.89}\) & \(36.5^{0.36}\) & \(41.6^{0.21}\) & \(36.2^{0.11}\) \\
& & & & \\
LWS & \(28.7\) & \(39.1\) & \(40.6\) & \(37.6\) \\
\(\alpha\)‑LWS & & & & \\
\(\rho=0.5\) & \(37.1^{1.39}\) & \(34.4^{0.80}\) & \(37.7^{0.52}\) & \(36.1^{0.31}\) \\
\(\rho=1\) & \(34.6^{0.97}\) & \(35.8^{0.54}\) & \(38.6^{0.39}\) & \(36.6^{0.22}\) \\
\(\rho=1.5\) & \(32.2^{1.17}\) & \(37.2^{0.36}\) & \(39.5^{0.39}\) & \(37.0^{0.11}\) \\
\bottomrule\noalign{}
\end{tabular}
\caption{\label{tbl:datasets_baselines_split_accs_vs_rho}Mean split accuracy in percents (standard deviation in superscript) of AlphaNet and various baseline methods on ImageNet‑LT and Places‑LT. \(\alpha\)‑cRT and \(\alpha\)‑LWS are AlphaNet models applied over cRT and LWS features respectively.}
\end{table}

\textbf{Baseline models.} First, we applied AlphaNet to models fine-tuned using classifier re-training (cRT), and learnable weight scaling (LWS)\citep{2020.Kalantidis.Kang}. These models have good overall accuracy, but accuracy for `few' split classes is much lower. On ImageNet‑LT, average `few' split accuracy using a ResNeXt‑50 backbone is around 20 points below the overall accuracy for both cRT and LWS, as seen in Table~\ref{tbl:datasets_baselines_split_accs_vs_rho}, which also shows other baseline methods -- nearest classifier mean (NCM), which predicts the nearest neighbor using average class representation, and \(\tau\)‑normalized, which scales classifier weights by their \(\tau\)-norm\citep{2020.Kalantidis.Kang}.

Using features extracted from the cRT and LWS models, we used AlphaNet to update `few' split classifiers creating \(\alpha\)‑cRT and \(\alpha\)‑LWS respectively. Per-split accuracies, obtained by training with \(\rho\) in \(\set{0.5, 1, 1.5}\), are shown in Table~\ref{tbl:datasets_baselines_split_accs_vs_rho}. We get a significant increase in the `few' split accuracy for all values of \(\rho\). Moreover, we see that \(\rho\) allows us to control the balance between `few' split and overall accuracies. Using larger values of \(\rho\) -- i.e., training with more `base' split samples -- allows overall accuracy to remain closer to the original, while still affording significant gains to `few' split accuracy. With \(\rho=1.5\), \(\alpha\)‑cRT boosts `few' split accuracy by more than 5 points, while overall accuracy is within about 1 point. \(\alpha\)‑LWS achieves even larger gains, increasing `few' split accuracy to around 40\%, while still maintaining a competitive 48\% overall accuracy.

We repeated the above experiment on Places‑LT, where we see similar performance gains on the `few' split (Table~\ref{tbl:datasets_baselines_split_accs_vs_rho}). Notably, with \(\rho=1\), \(\alpha\)‑LWS increases `few' split accuracy by about 6 points, while overall accuracy is within 1 point of the LWS model.

\hypertarget{tbl:datasets_split_accs_vs_rho_ride}{}
\begin{table}
\centering
\begin{tabular}[]{@{}lrrrr@{}}

\toprule\noalign{}
Method & Few & Med. & Many & Overall \\
\midrule\noalign{}

\textbf{ImageNet‑LT} & & & & \\
RIDE & \(36.5\) & \(54.4\) & \(68.9\) & \(57.5\) \\
\(\alpha\)‑RIDE & & & & \\
\(\rho=0.5\) & \(43.5^{0.75}\) & \(52.3^{0.26}\) & \(67.3^{0.17}\) & \(56.9^{0.11}\) \\
\(\rho=1\) & \(40.8^{1.00}\) & \(53.1^{0.21}\) & \(67.9^{0.18}\) & \(57.1^{0.11}\) \\
\(\rho=1.5\) & \(38.2^{1.22}\) & \(53.6^{0.25}\) & \(68.4^{0.17}\) & \(57.2^{0.06}\) \\
& & & & \\
\textbf{CIFAR‑100‑LT} & & & & \\
RIDE & \(25.8\) & \(52.1\) & \(69.3\) & \(50.2\) \\
\(\alpha\)‑RIDE & & & & \\
\(\rho=0.5\) & \(32.3^{1.24}\) & \(45.9^{0.87}\) & \(64.6^{0.78}\) & \(48.4^{0.43}\) \\
\(\rho=1\) & \(27.6^{1.41}\) & \(49.5^{0.83}\) & \(67.4^{0.70}\) & \(49.2^{0.16}\) \\
\(\rho=1.5\) & \(25.2^{1.11}\) & \(50.2^{0.57}\) & \(68.3^{0.34}\) & \(49.0^{0.26}\) \\
\bottomrule\noalign{}
\end{tabular}
\caption{\label{tbl:datasets_split_accs_vs_rho_ride}Mean split accuracy in percents (standard deviation in superscript) on ImageNet‑LT and CIFAR‑100‑LT using the ensemble RIDE model. \(\alpha\)‑RIDE applies AlphaNet on average features from the ensemble.}
\end{table}

\hypertarget{tbl:datasets_split_accs_vs_rho_ltr}{}
\begin{table}
\centering
\begin{tabular}[]{@{}lrrrr@{}}

\toprule\noalign{}
Method & Few & Med. & Many & Overall \\
\midrule\noalign{}

\textbf{CIFAR‑100‑LT} & & & & \\
LTR & \(29.8\) & \(49.3\) & \(70.1\) & \(50.7\) \\
\(\alpha\)‑LTR & & & & \\
\(\rho=0.5\) & \(36.2^{1.39}\) & \(39.5^{2.11}\) & \(63.4^{4.18}\) & \(46.9^{1.95}\) \\
\(\rho=1\) & \(32.2^{1.77}\) & \(43.4^{1.22}\) & \(67.5^{0.95}\) & \(48.5^{0.73}\) \\
\(\rho=1.5\) & \(32.0^{1.49}\) & \(46.2^{0.86}\) & \(67.3^{1.02}\) & \(49.3^{0.33}\) \\
\bottomrule\noalign{}
\end{tabular}
\caption{\label{tbl:datasets_split_accs_vs_rho_ltr}Mean split accuracy in percents (standard deviation in superscript) on CIFAR‑100‑LT using the LTR model.}
\end{table}

\textbf{State-of-the-art models.} Next, we applied AlphaNet to two state-of-the-art models: (1) the 6-expert ensemble RIDE model\citep{2021.Yu.Wang}, and (2) the weight balancing LTR model\citep{2022.Kong.Alshammari}. See Section~\ref{sec:impl:baselines:extract} for details on feature extraction for these models. Table~\ref{tbl:datasets_split_accs_vs_rho_ride} shows the base results for RIDE, along with AlphaNet results for \(\rho \in \set{0.5, 1, 1.5}\). On ImageNet‑LT, `few' split accuracy was increased by up to 7 points, and on CIFAR‑100‑LT, by 5 points. For the LTR model, we show results on CIFAR‑100‑LT in Table~\ref{tbl:datasets_split_accs_vs_rho_ltr} -- we are able to increase `few' split accuracy by almost 7 points.

These results show that AlphaNet can be applied reliably with state-of-the-art models to significantly improve the accuracy for rare classes.

\hypertarget{sec:exp:control}{%
\subsection{Comparison with control}\label{sec:exp:control}}

\begin{figure}
\hypertarget{fig:euclidean_random_split_deltas_vs_imagenetlt_crt}{%
\centering
\includegraphics{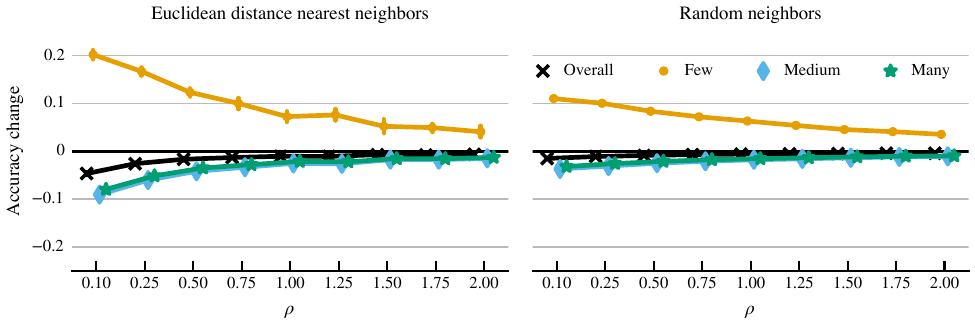}
\caption{Change in split accuracies for \(\alpha\)‑cRT on ImageNet‑LT. For each value of \(\rho\), the two plots show the raw difference in split accuracy (with accuracy expressed as a fraction) for AlphaNet compared to the baseline cRT model. Left shows the results for normal training with 5 nearest neighbors by Euclidean distance, and right shows the results for training with 5 random ``neighbors'' for each `few' split class. Training with nearest neighbors leads to a larger increase in `few' split accuracy, especially for small \(\rho\), which cannot be accounted for by the additional fine-tuning of classifiers alone.}\label{fig:euclidean_random_split_deltas_vs_imagenetlt_crt}
}
\end{figure}

Our method is based on the core hypothesis that classifiers can be improved using nearest neighbors. In this section, we directly evaluate this hypothesis. Based on the results in the previous section, the improvements in `few' split accuracy could be attributed simply to the extra fine-tuning of the classifiers. So, using the cRT model on ImageNet‑LT, we retrained AlphaNet with 5 randomly chosen `base' split classes as ``neighbors'' for each `few' split class. This differs from our previous experiments only in the classes used to update `few' split classifiers, so if AlphaNet's improvements were solely due to extra fine-tuning, we should see similar results. However, as seen in Figure~\ref{fig:euclidean_random_split_deltas_vs_imagenetlt_crt}, training with nearest neighbors selected by Euclidean distance garners much larger improvements in `few' split accuracy, with similar trends in overall accuracy. This supports our hypothesis that classifiers for data-poor classes can make use of information from visually similar classes to improve classification performance.

\hypertarget{sec:exp:predchanges}{%
\subsection{Prediction changes}\label{sec:exp:predchanges}}

\begin{figure}
\hypertarget{fig:cls_delta_vs_nndist}{%
\centering
\includegraphics{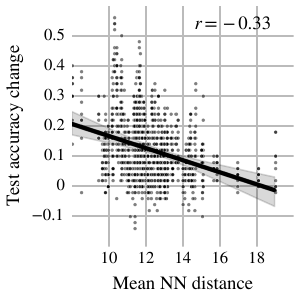}
\caption{Change in per-class test accuracy for `few' split of ImageNet‑LT with \(\alpha\)‑cRT, versus mean Euclidean distance to 5 nearest neighbors. Comparing with Figure~\ref{fig:analysis:acc_vs_dist}, we see that AlphaNet provides the largest boost to classes with close nearest neighbors, which have poor baseline performance.}\label{fig:cls_delta_vs_nndist}
}
\end{figure}

\begin{figure}
\subfloat[Predictions on `few' split classes, with NNs selected from the `base' split.]{\includegraphics{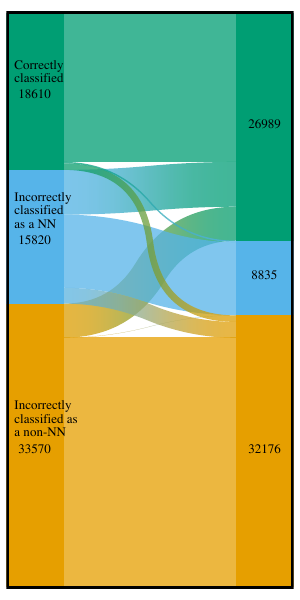}\label{fig:pred_changes:few}}
\quad
\subfloat[Predictions on `base' split classes, with NNs selected from the `few' split.]{\includegraphics{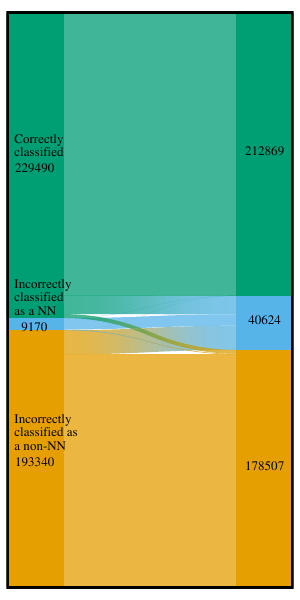}\label{fig:pred_changes:base}}
\quad
\subfloat[All predictions, with NNs selected from all classes. The hatched portions represent the `few' split.]{\includegraphics{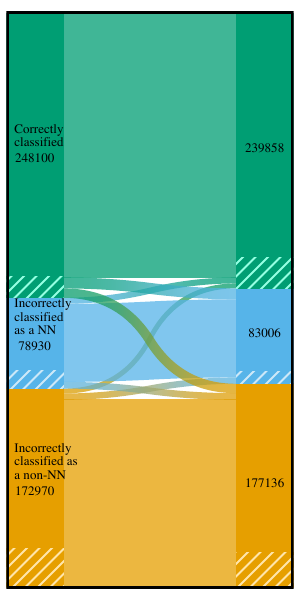}\label{fig:pred_changes:all}}
\caption{Change in sample predictions for \(\alpha\)‑cRT (\(\rho=0.5\)) on ImageNet‑LT. For each plot, the bars on the left show the distribution of predictions by the baseline model; and the bars on the right show the distribution for \(\alpha\)‑cRT. The groupings follow the scheme described in Figure~\ref{fig:analysis:bins}. The counts are aggregated from 10 repetitions of training \(\alpha\)‑cRT. The ``flow'' bands from left to right show the changes in individual sample predictions.}
\label{fig:pred_changes}
\end{figure}

As shown in Section~\ref{sec:intro}, the cRT model frequently misclassifies `few' split classes as visually similar `base' split classes. Using the AlphaNet model with \(\rho=0.5\), we performed the same analyses as before. Figure~\ref{fig:pred_changes:few} shows the change in sample predictions, where we see that a large portion of samples previously misclassified as a nearest neighbor are correctly classified after their classifiers are updated with AlphaNet. Furthermore, as seen in Figure~\ref{fig:cls_delta_vs_nndist}, AlphaNet improvements are strongly correlated to mean nearest neighbor distance. Classes with close neighbors, which had a high likelihood of being misclassified by the baseline model, see the biggest improvement in test accuracy.

\begin{figure}
\subfloat[Predictions on `few' split classes.]{\includegraphics{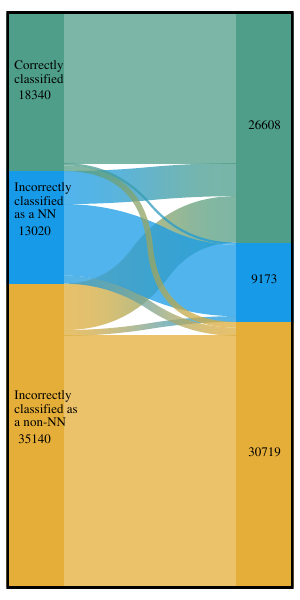}\label{fig:pred_changes_semantic:few}}
\quad
\subfloat[Predictions on `base' split classes.]{\includegraphics{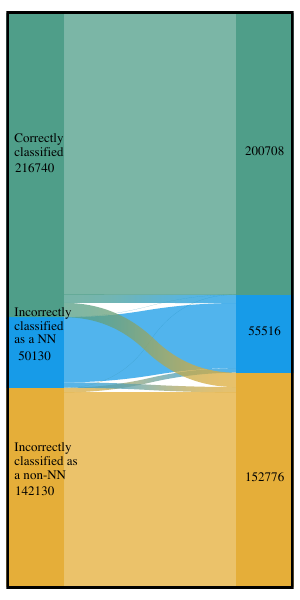}\label{fig:pred_changes_semantic:base}}
\quad
\subfloat[All predictions; hatched portions represent `few' split.]{\includegraphics{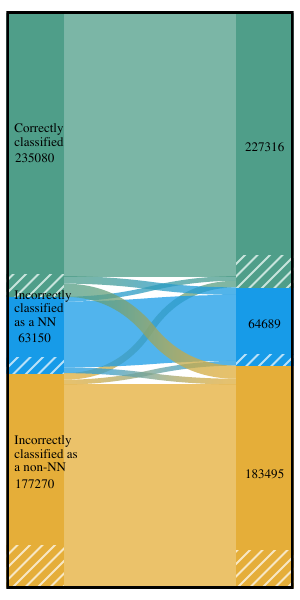}\label{fig:pred_changes_semantic:all}}
\caption{Change in sample predictions for \(\alpha\)‑cRT (\(\rho=0.5\)) grouped with respect to nearest neighbors identified using WordNet. This figure shows the same results as Figure~\ref{fig:pred_changes}, but grouped using differently defined nearest neighbors -- the new nearest neighbors are only used for visualizing.}
\label{fig:pred_changes_semantic}
\end{figure}

\hypertarget{sec:exp:analysis}{%
\subsection{Analysis of AlphaNet predictions}\label{sec:exp:analysis}}

AlphaNet significantly boosts the accuracy of `few' split classes. However, we do see a decrease in overall accuracy compared to baseline models, particularly for small values of \(\rho\). It is important to note that the increase in `few' split accuracy is much larger than the decrease in overall accuracy. As discussed earlier, in many applications it is important to have balanced performance across classes, and AlphaNet succeeds in making accuracies more balanced across splits.

However, we further analyzed the prediction changes for `base' split samples. Specifically, Figure~\ref{fig:pred_changes:base} shows change in predictions for `base' split samples, with nearest neighbors selected from the `few' split. We see a small increase in misclassifications as `few' split classes. This leads to the slight decrease in overall accuracy, which is also evident in Figure~\ref{fig:pred_changes:all} where all predictions are shown, and with nearest neighbors from all classes.

This previous analysis was conducted using nearest neighbors identified based on visual similarity. Since this is dependent on the representation space of the particular model, we conducted an additional analysis to see the behavior of predictions with respect to \emph{semantically similar} categories. For classes in ImageNet‑LT, we defined nearest neighbors using distance in the WordNet\citep{2010.Princeton} hierarchy. Specifically, if two classes (e.g., `Lhasa' and `Tibetan terrier') share a parent at most 4 levels higher in WordNet (in this example, `dog'), we consider them to be one of each other's nearest neighbors. Figure~\ref{fig:pred_changes_semantic} shows \(\alpha\)‑cRT predictions grouped using nearest neighbors defined his way. We see that a large number of incorrect predictions are to semantically similar categories which can be hard for even humans to distinguish. This suggests that metrics for long-tail classification should be re-evaluated for large datasets with many similar classes. Considering only misclassifications to semantically dissimilar classes, we see that AlphaNet still improves performance on `few' split classes, while maintaining overall accuracy. So, despite model-specific visual similarity, AlphaNet garners improvements at the semantic level, showing that it can be applied to models beyond those used in this paper.

%% file: src/section_5_conclusion.tex
\hypertarget{sec:conclusion}{%
\section{Conclusion}\label{sec:conclusion}}

The long-tailed nature of the world presents a challenge for classification models, due to the imbalance in the number of training samples per class. To address this problem, a number of methods have been proposed, but the focus is generally on achieving the highest overall accuracy. Consequently, many long-tail methods tend to have high overall accuracy, but with unbalanced per-class accuracies where frequent classes are learned well with high accuracies, and rare classes are learned poorly with low accuracies. Such models can lead to biased outcomes, which raises serious ethical concerns. In this paper, we proposed AlphaNet, a rapid post hoc correction method that can be applied to any model. Our simple method greatly improves the accuracy for data-poor classes, and re-balances per-class classification accuracies while preserving overall accuracy. AlphaNet can be deployed in any application where the base classifiers cannot be changed, but balanced performance is desirable -- thereby making it useful in contexts where ethics, privacy, or intellectual property are concerns.

%% file: src/section_6_acknowledgements.tex
\hypertarget{acknowledgements}{%
\section*{Acknowledgements}\label{acknowledgements}}
\addcontentsline{toc}{section}{Acknowledgements}

This paper is based upon work supported by the Department of Defense contract FA8702-15-D-0002, and the NSF Graduate Research Fellowship for Nadine Chang.

%% file: src/appendix_implementation.tex
\hypertarget{sec:impl}{%
\section{Implementation details}\label{sec:impl}}

Experiments were run using the PyTorch\citep{2019.Chintala.Paszke} library. We used the container implementation provided by NVIDIA GPU cloud (NGC).\footnote{\href{https://catalog.ngc.nvidia.com/orgs/nvidia/containers/pytorch}{\texttt{catalog.ngc.nvidia.com/orgs/nvidia/containers/pytorch}}, version 22.06.} Code to reproduce experimental results is available on GitHub.\footnote{\href{https://github.com/jayanthkoushik/alphanet}{\texttt{github.com/jayanthkoushik/alphanet}}.}

\hypertarget{sec:impl:datasets}{%
\subsection{Datasets}\label{sec:impl:datasets}}

\hypertarget{tbl:dataset_stats}{}
\begin{table}
\centering
\begin{tabular}[]{@{}lccc@{}}

\toprule\noalign{}
Dataset & ~~~~~Samples & ~~~~~Classes & ~~~~~~~~~Train samples \\
\midrule\noalign{}

& \(\qquad\text{train}\qquad\text{val}\qquad\text{test}\) & \(\quad\ \text{total}\quad\text{many}\ \ \quad\text{med.}\quad\text{few}\) & \(\qquad\text{many}\ \  \quad\text{med.}\quad\ \ \text{few}\) \\
& & & \\
ImageNet‑LT & \(\qquad 115,846 \quad 20,000 \quad 50,000\) & \(\qquad 1,000 \quad 385 \quad\phantom{,0}479 \quad 136\phantom{,0}\) & \(\qquad\phantom{0}88,693 \quad 25,510 \quad 1,643\phantom{0}\) \\
Places‑LT & \(\qquad\phantom{0}62,500 \quad\phantom{0}7,300 \quad 36,500\) & \(\qquad 365\phantom{,0} \quad 131 \quad\phantom{,0}163 \quad 71\phantom{0,0}\) & \(\qquad\phantom{0}52,762 \quad\phantom{0}8,934 \quad 804\phantom{0,0}\) \\
CIFAR‑100‑LT & \(\qquad\phantom{0}10,847 \quad\phantom{00,000} \quad 10,000\) & \(\qquad 100\phantom{,0} \quad\phantom{0}35 \quad\phantom{0,0}35 \quad 30\phantom{0,0}\) & \(\qquad\phantom{00}8,824 \quad\phantom{0}1,718 \quad 305\phantom{0,0}\) \\
iNaturalist & \(\qquad 437,513 \quad\phantom{00,000} \quad 24,426\) & \(\qquad 8,142 \quad 842 \quad 4,076 \quad 3,224\) & \(\qquad 258,340 \quad 133,061 \quad 46,112\) \\
\bottomrule\noalign{}
\end{tabular}
\caption{\label{tbl:dataset_stats}Statistics of long-tailed datasets.}
\end{table}

Details about the long-tailed datasets used in our experiments are shown in Table~\ref{tbl:dataset_stats}. For ImageNet‑LT and Places‑LT, we used splits from Kang et~al.\citep{2020.Kalantidis.Kang}, available on GitHub.\footnote{\href{https://github.com/facebookresearch/classifier-balancing}{\texttt{github.com/facebookresearch/classifier-balancing}}.} For CIFAR‑100‑LT, we used the implementation of Wang et~al.\citep{2021.Yu.Wang}, with imbalance factor 100, also available on GitHub.\footnote{\href{https://github.com/frank-xwang/RIDE-LongTailRecognition}{\texttt{github.com/frank-xwang/RIDE-LongTailRecognition}}.}

\hypertarget{sec:impl:datasets:splits}{%
\subsubsection{Splits}\label{sec:impl:datasets:splits}}

For all datasets, the `many', `medium', and `few' splits are defined using the same limits on per-class training samples: less than 20 for the `few' split, between 20 and 100 for the `medium' split, and more than 100 for the `many' split. The actual minimum and maximum per-class training samples for each split are shown in Table~\ref{tbl:dataset_splits}.

\hypertarget{sec:impl:baselines}{%
\subsection{Baseline models}\label{sec:impl:baselines}}

Baseline model architectures are shown in Table~\ref{tbl:baseline_archs}. All models used backbones made of residual networks -- ResNets\citep{2016.Sun.He}, and ResNeXts\citep{2017.He.Xie}. Whenever we refer to a model, the architecture corresponding to the dataset is used. For example, cRT used the ResNeXt‑50 architecture on ImageNet‑LT, and the ResNet‑152 architecture on Places‑LT.

For all models except LTR, we used model weights provided by the respective authors. For LTR, we retrained the model using code provided by the authors,\footnote{\href{https://github.com/ShadeAlsha/LTR-weight-balancing}{\texttt{github.com/ShadeAlsha/LTR-weight-balancing}}.} with some modifications: (1) for consistency, we used the same CIFAR‑100‑LT data splits used for training the RIDE model, and (2) we performed second stage training -- fine-tuning with weight decay and norm thresholding -- for a fixed 10 epochs.

\hypertarget{tbl:dataset_splits}{}
\begin{table}
\centering
\begin{tabular}[]{@{}lcc@{}}

\toprule\noalign{}
Dataset & Min. per-class samples & Max. per-class samples \\
\midrule\noalign{}

& \(\ \ \text{many}\ \  \text{med.}\ \ \text{few}\ \) & \(\ \ \ \text{many}\ \ \  \text{med.}\ \ \ \text{few}\ \) \\
& & \\
ImageNet‑LT & \(\  101 \quad 20 \quad 5\ \) & \(1,280 \quad 100 \quad 19\ \) \\
Places‑LT & \(\  103 \quad 20 \quad 5\ \) & \(4,980 \quad 100 \quad 19\ \) \\
CIFAR‑100‑LT & \(\  102 \quad 20 \quad 5\ \) & \(\phantom{,0}500 \quad\phantom{0}98 \quad 19\ \) \\
iNaturalist & \(\  101 \quad 20 \quad 2\ \) & \(1,000 \quad 100 \quad 19\ \) \\
\bottomrule\noalign{}
\end{tabular}
\caption{\label{tbl:dataset_splits}Minimum and maximum per-class training samples for long-tailed datasets.}
\end{table}

\hypertarget{sec:impl:baselines:extract}{%
\subsubsection{Feature and classifier extraction}\label{sec:impl:baselines:extract}}

\hypertarget{tbl:baseline_archs}{}
\begin{table}
\centering
\begin{tabular}[]{@{}lll@{}}

\toprule\noalign{}
Dataset & Model & Architecture \\
\midrule\noalign{}

ImageNet‑LT & cRT & ResNeXt‑50 \\
& LWS & ResNeXt‑50 \\
& RIDE & ResNeXt‑50 \\
& & \\
Places‑LT & cRT & ResNet‑152 \\
& LWS & ResNet‑152 \\
& & \\
CIFAR‑100‑LT & RIDE & ResNet‑32 \\
& LTR & ResNet‑34 \\
& & \\
iNaturalist & cRT & ResNet‑152 \\
\bottomrule\noalign{}
\end{tabular}
\caption{\label{tbl:baseline_archs}Baseline model architectures.}
\end{table}

In most cases, we simply used the output of a model's penultimate layer as features, and used the weights (including bias) of the last layer as the classifier. Exceptions are listed below:

\begin{itemize}
\item
  LWS: We multiplied classifier weights with the learned scales.
\item
  RIDE: We used the 6-expert teacher model, and saved classifiers from each expert after normalizing and scaling as in the model. For AlphaNet training, we created a single classifier by concatenating the expert classifier weights and biases. Similarly, features extracted from individual experts were concatenated after normalizing. During prediction, the individual experts and features were re-extracted, and experts were applied to their corresponding features to get 6 predictions, which were then averaged. So AlphaNet learned coefficients to update all 6 experts simultaneously.
\item
  LTR: We used the model fine-tuned with weight decay and norm thresholding. This creates a classifier with small norm, and correspondingly small prediction scores. So, during AlphaNet training, we multiplied all prediction scores by 100, which is equivalent to setting the softmax temperature to 0.01.
\end{itemize}

\hypertarget{sec:impl:training}{%
\subsection{Training}\label{sec:impl:training}}

For the main experiments, 5 nearest neighbors were selected for each `few' split class, based on Euclidean distance. Hyper-parameter settings used during training are shown in Table~\ref{tbl:hyperparams}. ImageNet‑LT and Places‑LT have a validation set, which was used to select the best model. This was controlled by the `minimum epochs' parameter. After training for at least this many epochs, model weights were saved at the end of each epoch. Finally, the best model was selected based on overall validation accuracy, and used for testing. For CIFAR‑100‑LT and iNaturalist, we simply trained for a fixed number of epochs.

\hypertarget{tbl:hyperparams}{}
\begin{table}
\centering
\begin{tabular}[]{@{}ll@{}}

\toprule\noalign{}
Parameter & Value \\
\midrule\noalign{}

Optimizer & AdamW\citep{2019.Hutter.Loshchilov} with default parameters \\
& (\(\beta_1=0.9, \beta_2=0.999, \epsilon=0.01, \lambda=0.01\)) \\
Initial learning rate & \(0.001\) \\
Learning rate decay & \(0.1\) every 10 epochs \\
Training epochs & 10 (CIFAR‑100‑LT and iNaturalist) \\
& 25 (ImageNet‑LT and Places‑LT) \\
Minimum epochs & 5 \\
Batch size & 256 for iNaturalist, 64 for all others \\
AlphaNet architecture & 3 fully connected layers each with 32 units \\
Hidden layer activation & Leaky-ReLU\citep{2011.Bengio.Glorot} with negative slope 0.01 \\
Weight initialization & Uniform sampling with bounds \(\pm 1/\sqrt{m}\) \\
& where \(m\) is the number of input units to a layer \\
\bottomrule\noalign{}
\end{tabular}
\caption{\label{tbl:hyperparams}Training hyper-parameters for main experiments.}
\end{table}

\hypertarget{sec:impl:results}{%
\subsection{Results}\label{sec:impl:results}}

All experiments were repeated 10 times from different random initializations, and unless specified otherwise, results are average values. In tables, the standard deviation is shown in superscript. We regenerated baseline results, and these match published values, except in the case of LTR, since it was retrained, and, for consistency, we did not use data augmentation at test time. Plots were generated with Matplotlib\citep{2007.Hunter}, using the Seaborn library\citep{2021.Waskom}. Error bars in figures represent 95\% confidence intervals, estimated using 10,000 bootstrap resamples.

%% file: src/appendix_ksweep.tex
\graphicspath{{figures/appendix/}}

\hypertarget{sec:ksweep}{%
\section{Analysis of nearest neighbor selection}\label{sec:ksweep}}

We analyzed the effect of the number of nearest neighbors \(k\), and the distance metric (\(\mu\)), on the performance of AlphaNet, using the cRT model on ImageNet‑LT. We compared two distance metrics:

\begin{itemize}
\tightlist
\item
  Cosine distance: \(\mu(z_1, z_2) = 1 - z_1^T z_2\).
\item
  Euclidean distance: \(\mu(z_1, z_2) = \left\|{z_1 - z_2}\right\|_2\).
\end{itemize}

For each distance metric, we performed 4 sets of experiments, with \(\rho\) in \(\left\{0.25, 0.5, 1, 2\right\}\). For each \(\rho\), we varied \(k\) from 2 to 10; all other hyper-parameters were kept the same as described in Section~\ref{sec:impl}.

\begin{figure}
\subfloat[\(\rho=0.25\)]{\includegraphics{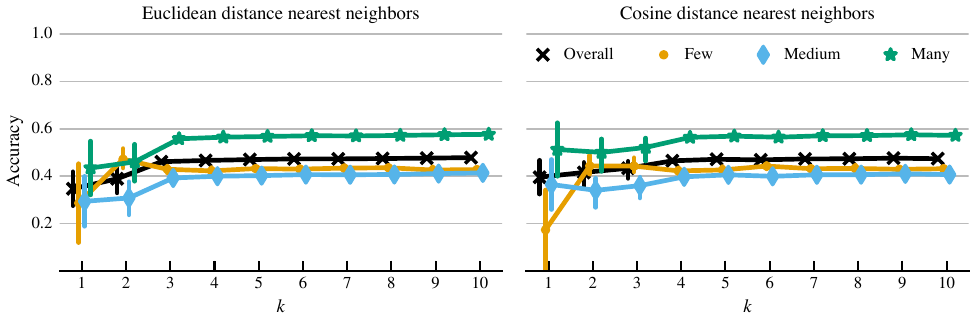}\label{fig:ksweep:025}}
\quad
\subfloat[\(\rho=0.5\)]{\includegraphics{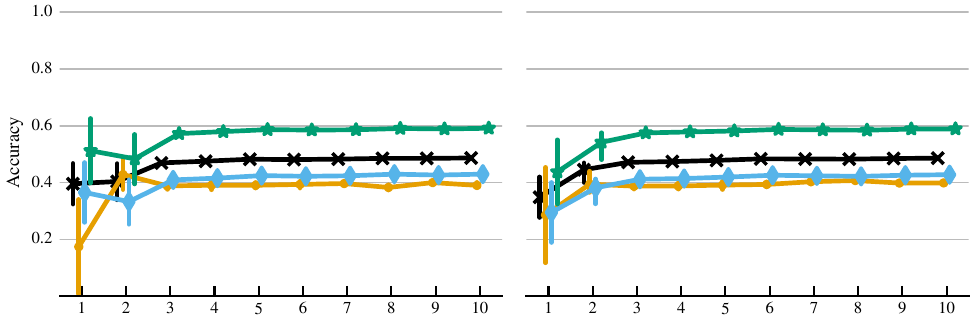}\label{fig:ksweep:05}}
\quad
\subfloat[\(\rho=1\)]{\includegraphics{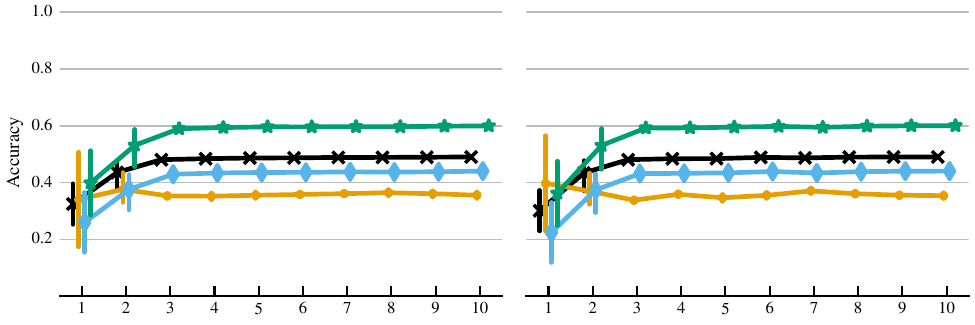}\label{fig:ksweep:1}}
\caption{Per-split test accuracies for \(\alpha\)‑cRT on ImageNet‑LT versus the number of nearest neighbors \(k\).}
\label{fig:ksweep}
\end{figure}

The results are summarized in Figure~\ref{fig:ksweep}, which shows per-split top‑1 accuracies against \(k\) for different values of \(\rho\) (\(\rho=2\) is omitted from this figure for space -- no special behavior was observed for this case). We observe little change in performance beyond \(k=5\), and also observe similar performance for both distance metrics.

The full set of top‑1 and top‑5 accuracies is shown in the following tables:

\begin{itemize}
\tightlist
\item
  Euclidean distance:

  \begin{itemize}
  \tightlist
  \item
    Top‑1 accuracy: Table~\ref{tbl:rhos_split_top1_accs_vs_k_imagenetlt_crt_euclidean}.
  \item
    Top‑5 accuracy: Table~\ref{tbl:rhos_split_top5_accs_vs_k_imagenetlt_crt_euclidean}.
  \end{itemize}
\item
  Cosine distance:

  \begin{itemize}
  \tightlist
  \item
    Top‑1 accuracy: Table~\ref{tbl:rhos_split_top1_accs_vs_k_imagenetlt_crt_cosine}.
  \item
    Top‑5 accuracy: Table~\ref{tbl:rhos_split_top5_accs_vs_k_imagenetlt_crt_cosine}.
  \end{itemize}
\end{itemize}

\clearpage

\clearpage

\hypertarget{tbl:rhos_split_top1_accs_vs_k_imagenetlt_crt_euclidean}{}
\begin{table}
\centering
\begin{tabular}[]{@{}lrrrr@{}}

\toprule\noalign{}
Model & Few & Med. & Many & Overall \\
\midrule\noalign{}

cRT & \(27.4\) & \(46.2\) & \(61.8\) & \(49.6\) \\
& & & & \\
\(\bm{\alpha}\)\textbf{‑cRT} & & & & \\
\(\rho=0.25\) & & & & \\
\(k=1\) & \(28.7^{29.06}\) & \(29.4^{18.36}\) & \(43.5^{19.74}\) & \(34.7^{12.44}\) \\
\(k=2\) & \(47.2^{07.24}\) & \(30.9^{11.45}\) & \(46.1^{13.08}\) & \(38.9^{09.54}\) \\
\(k=3\) & \(43.0^{02.40}\) & \(39.3^{00.63}\) & \(55.8^{00.48}\) & \(46.1^{00.37}\) \\
\(k=4\) & \(42.3^{02.36}\) & \(40.0^{01.00}\) & \(56.5^{00.79}\) & \(46.7^{00.51}\) \\
\(k=5\) & \(43.2^{01.58}\) & \(40.3^{00.57}\) & \(56.8^{00.48}\) & \(47.0^{00.38}\) \\
\(k=6\) & \(43.0^{01.81}\) & \(40.7^{01.01}\) & \(57.1^{00.84}\) & \(47.3^{00.56}\) \\
\(k=7\) & \(43.5^{01.95}\) & \(40.6^{00.81}\) & \(57.0^{00.67}\) & \(47.3^{00.43}\) \\
\(k=8\) & \(43.6^{01.47}\) & \(40.7^{00.58}\) & \(57.2^{00.50}\) & \(47.5^{00.34}\) \\
\(k=9\) & \(42.6^{02.26}\) & \(41.1^{00.88}\) & \(57.5^{00.76}\) & \(47.6^{00.43}\) \\
\(k=10\) & \(42.8^{01.01}\) & \(41.4^{00.58}\) & \(57.7^{00.41}\) & \(47.9^{00.32}\) \\
& & & & \\
\(\rho=0.5\) & & & & \\
\(k=1\) & \(17.5^{26.76}\) & \(36.5^{16.92}\) & \(51.2^{18.25}\) & \(39.6^{11.49}\) \\
\(k=2\) & \(42.8^{09.53}\) & \(33.3^{12.77}\) & \(48.4^{14.22}\) & \(40.4^{10.30}\) \\
\(k=3\) & \(38.9^{02.06}\) & \(41.0^{00.45}\) & \(57.2^{00.60}\) & \(46.9^{00.40}\) \\
\(k=4\) & \(39.1^{02.53}\) & \(41.6^{00.90}\) & \(57.9^{00.64}\) & \(47.5^{00.56}\) \\
\(k=5\) & \(39.1^{01.64}\) & \(42.5^{00.52}\) & \(58.6^{00.39}\) & \(48.2^{00.19}\) \\
\(k=6\) & \(39.4^{02.05}\) & \(42.2^{00.51}\) & \(58.5^{00.44}\) & \(48.1^{00.21}\) \\
\(k=7\) & \(39.7^{01.21}\) & \(42.4^{00.51}\) & \(58.6^{00.44}\) & \(48.3^{00.29}\) \\
\(k=8\) & \(38.3^{01.54}\) & \(43.0^{00.55}\) & \(59.0^{00.39}\) & \(48.5^{00.26}\) \\
\(k=9\) & \(40.0^{00.99}\) & \(42.7^{00.39}\) & \(58.9^{00.33}\) & \(48.5^{00.21}\) \\
\(k=10\) & \(39.0^{01.64}\) & \(43.0^{00.51}\) & \(59.1^{00.41}\) & \(48.7^{00.21}\) \\
& & & & \\
\(\rho=1\) & & & & \\
\(k=1\) & \(34.1^{28.61}\) & \(26.0^{18.09}\) & \(39.9^{19.51}\) & \(32.5^{12.28}\) \\
\(k=2\) & \(37.7^{08.93}\) & \(37.6^{10.62}\) & \(53.1^{11.56}\) & \(43.6^{08.36}\) \\
\(k=3\) & \(35.4^{01.48}\) & \(42.9^{00.50}\) & \(59.0^{00.46}\) & \(48.1^{00.25}\) \\
\(k=4\) & \(35.2^{02.02}\) & \(43.4^{00.53}\) & \(59.4^{00.46}\) & \(48.4^{00.30}\) \\
\(k=5\) & \(35.6^{01.93}\) & \(43.6^{00.59}\) & \(59.6^{00.41}\) & \(48.7^{00.23}\) \\
\(k=6\) & \(35.8^{01.23}\) & \(43.6^{00.39}\) & \(59.6^{00.32}\) & \(48.7^{00.24}\) \\
\(k=7\) & \(36.1^{01.30}\) & \(43.8^{00.45}\) & \(59.7^{00.30}\) & \(48.9^{00.17}\) \\
\(k=8\) & \(36.5^{01.90}\) & \(43.7^{00.43}\) & \(59.7^{00.36}\) & \(48.9^{00.16}\) \\
\(k=9\) & \(36.1^{01.80}\) & \(43.8^{00.45}\) & \(59.8^{00.40}\) & \(48.9^{00.15}\) \\
\(k=10\) & \(35.6^{02.01}\) & \(44.0^{00.56}\) & \(60.0^{00.48}\) & \(49.0^{00.22}\) \\
& & & & \\
\(\rho=2\) & & & & \\
\(k=1\) & \(23.0^{28.61}\) & \(33.0^{18.09}\) & \(47.4^{19.51}\) & \(37.2^{12.28}\) \\
\(k=2\) & \(29.9^{01.82}\) & \(43.8^{00.45}\) & \(59.6^{00.46}\) & \(48.0^{00.16}\) \\
\(k=3\) & \(30.9^{01.86}\) & \(44.2^{00.35}\) & \(60.1^{00.29}\) & \(48.5^{00.16}\) \\
\(k=4\) & \(31.2^{02.05}\) & \(44.5^{00.40}\) & \(60.3^{00.39}\) & \(48.8^{00.21}\) \\
\(k=5\) & \(33.1^{01.45}\) & \(44.3^{00.31}\) & \(60.2^{00.32}\) & \(48.9^{00.15}\) \\
\(k=6\) & \(32.0^{01.68}\) & \(44.7^{00.34}\) & \(60.5^{00.21}\) & \(49.1^{00.07}\) \\
\(k=7\) & \(32.3^{01.38}\) & \(44.8^{00.31}\) & \(60.6^{00.25}\) & \(49.2^{00.14}\) \\
\(k=8\) & \(32.2^{01.70}\) & \(44.8^{00.38}\) & \(60.6^{00.28}\) & \(49.2^{00.11}\) \\
\(k=9\) & \(32.4^{01.22}\) & \(44.8^{00.29}\) & \(60.6^{00.18}\) & \(49.2^{00.07}\) \\
\(k=10\) & \(32.4^{01.85}\) & \(44.9^{00.47}\) & \(60.7^{00.37}\) & \(49.3^{00.14}\) \\
\bottomrule\noalign{}
\end{tabular}
\caption{\label{tbl:rhos_split_top1_accs_vs_k_imagenetlt_crt_euclidean}Per-split test top‑1 accuracies for \(\alpha\)‑cRT on ImageNet‑LT using \(k\) nearest neighbors based on Euclidean distance.}
\end{table}

\hypertarget{tbl:rhos_split_top5_accs_vs_k_imagenetlt_crt_euclidean}{}
\begin{table}
\centering
\begin{tabular}[]{@{}lrrrr@{}}

\toprule\noalign{}
Model & Few & Med. & Many & Overall \\
\midrule\noalign{}

cRT & \(57.3\) & \(73.4\) & \(81.8\) & \(74.4\) \\
& & & & \\
\(\bm{\alpha}\)\textbf{‑cRT} & & & & \\
\(\rho=0.25\) & & & & \\
\(k=1\) & \(42.5^{41.12}\) & \(63.7^{11.28}\) & \(74.8^{8.11}\) & \(65.1^{2.93}\) \\
\(k=2\) & \(71.4^{07.44}\) & \(65.5^{07.09}\) & \(76.1^{5.04}\) & \(70.4^{4.33}\) \\
\(k=3\) & \(67.3^{01.26}\) & \(70.4^{00.32}\) & \(79.6^{0.21}\) & \(73.5^{0.22}\) \\
\(k=4\) & \(67.8^{01.39}\) & \(70.5^{00.35}\) & \(79.7^{0.26}\) & \(73.7^{0.14}\) \\
\(k=5\) & \(68.2^{00.89}\) & \(70.7^{00.39}\) & \(79.8^{0.24}\) & \(73.9^{0.23}\) \\
\(k=6\) & \(68.1^{01.28}\) & \(70.8^{00.46}\) & \(79.9^{0.32}\) & \(73.9^{0.17}\) \\
\(k=7\) & \(68.9^{00.86}\) & \(70.7^{00.37}\) & \(79.8^{0.26}\) & \(73.9^{0.19}\) \\
\(k=8\) & \(68.7^{00.71}\) & \(70.8^{00.25}\) & \(79.8^{0.18}\) & \(74.0^{0.13}\) \\
\(k=9\) & \(68.4^{01.27}\) & \(70.9^{00.34}\) & \(79.9^{0.24}\) & \(74.0^{0.13}\) \\
\(k=10\) & \(68.5^{00.69}\) & \(71.0^{00.33}\) & \(79.9^{0.21}\) & \(74.1^{0.16}\) \\
& & & & \\
\(\rho=0.5\) & & & & \\
\(k=1\) & \(26.6^{37.90}\) & \(68.0^{10.37}\) & \(77.9^{7.45}\) & \(66.2^{2.68}\) \\
\(k=2\) & \(68.1^{09.42}\) & \(66.6^{07.36}\) & \(76.9^{5.22}\) & \(70.8^{4.26}\) \\
\(k=3\) & \(64.7^{01.62}\) & \(70.9^{00.31}\) & \(80.0^{0.26}\) & \(73.6^{0.18}\) \\
\(k=4\) & \(65.0^{02.06}\) & \(71.2^{00.51}\) & \(80.1^{0.30}\) & \(73.8^{0.32}\) \\
\(k=5\) & \(65.3^{00.78}\) & \(71.6^{00.17}\) & \(80.4^{0.14}\) & \(74.1^{0.08}\) \\
\(k=6\) & \(65.9^{01.05}\) & \(71.3^{00.21}\) & \(80.3^{0.18}\) & \(74.0^{0.12}\) \\
\(k=7\) & \(66.2^{00.93}\) & \(71.5^{00.34}\) & \(80.3^{0.24}\) & \(74.1^{0.16}\) \\
\(k=8\) & \(65.4^{00.94}\) & \(71.6^{00.30}\) & \(80.4^{0.20}\) & \(74.1^{0.13}\) \\
\(k=9\) & \(66.3^{00.71}\) & \(71.6^{00.17}\) & \(80.4^{0.13}\) & \(74.2^{0.09}\) \\
\(k=10\) & \(66.0^{00.90}\) & \(71.6^{00.25}\) & \(80.5^{0.16}\) & \(74.3^{0.10}\) \\
& & & & \\
\(\rho=1\) & & & & \\
\(k=1\) & \(50.1^{40.52}\) & \(61.6^{11.08}\) & \(73.3^{7.97}\) & \(64.5^{2.87}\) \\
\(k=2\) & \(63.4^{08.24}\) & \(69.0^{05.95}\) & \(78.6^{4.20}\) & \(71.9^{3.37}\) \\
\(k=3\) & \(61.9^{01.32}\) & \(71.8^{00.32}\) & \(80.6^{0.26}\) & \(73.8^{0.15}\) \\
\(k=4\) & \(62.0^{01.36}\) & \(72.0^{00.23}\) & \(80.7^{0.17}\) & \(74.0^{0.15}\) \\
\(k=5\) & \(62.7^{01.25}\) & \(72.0^{00.27}\) & \(80.8^{0.20}\) & \(74.1^{0.11}\) \\
\(k=6\) & \(63.2^{00.83}\) & \(72.0^{00.22}\) & \(80.8^{0.18}\) & \(74.2^{0.13}\) \\
\(k=7\) & \(63.5^{01.07}\) & \(72.0^{00.25}\) & \(80.7^{0.15}\) & \(74.2^{0.09}\) \\
\(k=8\) & \(64.2^{01.06}\) & \(72.0^{00.23}\) & \(80.7^{0.15}\) & \(74.3^{0.14}\) \\
\(k=9\) & \(63.7^{01.26}\) & \(72.1^{00.27}\) & \(80.8^{0.17}\) & \(74.3^{0.07}\) \\
\(k=10\) & \(63.6^{01.51}\) & \(72.1^{00.34}\) & \(80.8^{0.21}\) & \(74.3^{0.08}\) \\
& & & & \\
\(\rho=2\) & & & & \\
\(k=1\) & \(34.4^{40.52}\) & \(65.9^{11.08}\) & \(76.4^{7.97}\) & \(65.6^{2.87}\) \\
\(k=2\) & \(56.8^{01.38}\) & \(72.2^{00.30}\) & \(80.9^{0.23}\) & \(73.5^{0.16}\) \\
\(k=3\) & \(57.8^{01.48}\) & \(72.4^{00.19}\) & \(81.1^{0.20}\) & \(73.8^{0.17}\) \\
\(k=4\) & \(58.8^{01.32}\) & \(72.4^{00.25}\) & \(81.1^{0.19}\) & \(73.9^{0.18}\) \\
\(k=5\) & \(60.9^{01.15}\) & \(72.3^{00.27}\) & \(80.9^{0.18}\) & \(74.1^{0.12}\) \\
\(k=6\) & \(60.0^{01.31}\) & \(72.6^{00.17}\) & \(81.2^{0.13}\) & \(74.2^{0.09}\) \\
\(k=7\) & \(60.5^{00.97}\) & \(72.6^{00.19}\) & \(81.1^{0.15}\) & \(74.2^{0.09}\) \\
\(k=8\) & \(60.6^{01.15}\) & \(72.6^{00.21}\) & \(81.2^{0.20}\) & \(74.3^{0.06}\) \\
\(k=9\) & \(60.9^{00.99}\) & \(72.6^{00.19}\) & \(81.2^{0.13}\) & \(74.3^{0.04}\) \\
\(k=10\) & \(60.9^{01.34}\) & \(72.6^{00.23}\) & \(81.2^{0.16}\) & \(74.3^{0.08}\) \\
\bottomrule\noalign{}
\end{tabular}
\caption{\label{tbl:rhos_split_top5_accs_vs_k_imagenetlt_crt_euclidean}Per-split test top‑5 accuracies for \(\alpha\)‑cRT on ImageNet‑LT using \(k\) nearest neighbors based on Euclidean distance.}
\end{table}

\clearpage

\hypertarget{tbl:rhos_split_top1_accs_vs_k_imagenetlt_crt_cosine}{}
\begin{table}
\centering
\begin{tabular}[]{@{}lrrrr@{}}

\toprule\noalign{}
Model & Few & Med. & Many & Overall \\
\midrule\noalign{}

cRT & \(27.4\) & \(46.2\) & \(61.8\) & \(49.6\) \\
& & & & \\
\(\bm{\alpha}\)\textbf{‑cRT} & & & & \\
\(\rho=0.25\) & & & & \\
\(k=1\) & \(17.4^{26.99}\) & \(36.5^{16.97}\) & \(51.3^{18.15}\) & \(39.6^{11.45}\) \\
\(k=2\) & \(44.3^{06.71}\) & \(34.1^{10.34}\) & \(50.1^{11.57}\) & \(41.6^{08.50}\) \\
\(k=3\) & \(44.3^{05.19}\) & \(36.1^{08.08}\) & \(52.2^{09.06}\) & \(43.4^{06.69}\) \\
\(k=4\) & \(42.3^{02.12}\) & \(39.8^{00.74}\) & \(56.4^{00.70}\) & \(46.5^{00.52}\) \\
\(k=5\) & \(42.7^{02.81}\) & \(40.6^{01.04}\) & \(56.9^{00.93}\) & \(47.2^{00.52}\) \\
\(k=6\) & \(44.3^{01.89}\) & \(40.0^{00.98}\) & \(56.5^{00.81}\) & \(46.9^{00.54}\) \\
\(k=7\) & \(43.2^{01.60}\) & \(40.6^{00.97}\) & \(57.1^{00.77}\) & \(47.3^{00.55}\) \\
\(k=8\) & \(43.3^{02.31}\) & \(40.7^{01.01}\) & \(57.1^{00.86}\) & \(47.4^{00.52}\) \\
\(k=9\) & \(42.9^{01.23}\) & \(41.0^{00.72}\) & \(57.5^{00.62}\) & \(47.6^{00.45}\) \\
\(k=10\) & \(43.2^{01.54}\) & \(40.7^{00.59}\) & \(57.2^{00.44}\) & \(47.4^{00.32}\) \\
& & & & \\
\(\rho=0.5\) & & & & \\
\(k=1\) & \(28.6^{29.44}\) & \(29.5^{18.51}\) & \(43.8^{19.80}\) & \(34.9^{12.49}\) \\
\(k=2\) & \(39.5^{06.79}\) & \(38.2^{08.65}\) & \(54.2^{09.53}\) & \(44.5^{06.91}\) \\
\(k=3\) & \(38.8^{01.78}\) & \(41.2^{00.82}\) & \(57.5^{00.78}\) & \(47.1^{00.49}\) \\
\(k=4\) & \(38.8^{01.88}\) & \(41.4^{00.38}\) & \(57.8^{00.44}\) & \(47.4^{00.30}\) \\
\(k=5\) & \(39.1^{02.47}\) & \(42.0^{00.72}\) & \(58.2^{00.62}\) & \(47.8^{00.33}\) \\
\(k=6\) & \(39.4^{01.45}\) & \(42.6^{00.67}\) & \(58.7^{00.44}\) & \(48.4^{00.31}\) \\
\(k=7\) & \(40.3^{01.19}\) & \(42.4^{00.47}\) & \(58.5^{00.43}\) & \(48.3^{00.28}\) \\
\(k=8\) & \(40.7^{01.35}\) & \(42.2^{00.60}\) & \(58.4^{00.42}\) & \(48.3^{00.34}\) \\
\(k=9\) & \(39.8^{01.08}\) & \(42.6^{00.40}\) & \(58.8^{00.31}\) & \(48.5^{00.22}\) \\
\(k=10\) & \(39.9^{01.17}\) & \(42.8^{00.49}\) & \(58.9^{00.31}\) & \(48.6^{00.24}\) \\
& & & & \\
\(\rho=1\) & & & & \\
\(k=1\) & \(39.8^{26.99}\) & \(22.5^{16.97}\) & \(36.2^{18.15}\) & \(30.1^{11.45}\) \\
\(k=2\) & \(37.0^{09.44}\) & \(37.4^{11.58}\) & \(53.1^{12.34}\) & \(43.4^{09.03}\) \\
\(k=3\) & \(33.8^{01.49}\) & \(43.2^{00.52}\) & \(59.2^{00.48}\) & \(48.1^{00.25}\) \\
\(k=4\) & \(35.9^{01.10}\) & \(43.3^{00.38}\) & \(59.2^{00.24}\) & \(48.4^{00.27}\) \\
\(k=5\) & \(34.7^{01.99}\) & \(43.4^{00.52}\) & \(59.5^{00.39}\) & \(48.4^{00.22}\) \\
\(k=6\) & \(35.5^{01.79}\) & \(43.9^{00.41}\) & \(59.8^{00.42}\) & \(48.9^{00.14}\) \\
\(k=7\) & \(37.1^{01.72}\) & \(43.4^{00.48}\) & \(59.4^{00.43}\) & \(48.7^{00.22}\) \\
\(k=8\) & \(36.1^{01.47}\) & \(43.8^{00.40}\) & \(59.8^{00.31}\) & \(48.9^{00.15}\) \\
\(k=9\) & \(35.6^{01.37}\) & \(44.0^{00.32}\) & \(60.0^{00.30}\) & \(49.0^{00.14}\) \\
\(k=10\) & \(35.4^{01.68}\) & \(44.0^{00.51}\) & \(60.0^{00.33}\) & \(49.0^{00.15}\) \\
& & & & \\
\(\rho=2\) & & & & \\
\(k=1\) & \(23.0^{28.85}\) & \(33.0^{18.14}\) & \(47.5^{19.40}\) & \(37.2^{12.24}\) \\
\(k=2\) & \(29.1^{01.64}\) & \(44.0^{00.36}\) & \(59.9^{00.25}\) & \(48.1^{00.15}\) \\
\(k=3\) & \(30.9^{01.89}\) & \(44.1^{00.59}\) & \(60.0^{00.45}\) & \(48.4^{00.23}\) \\
\(k=4\) & \(31.6^{02.13}\) & \(44.3^{00.49}\) & \(60.3^{00.38}\) & \(48.7^{00.20}\) \\
\(k=5\) & \(32.5^{02.40}\) & \(44.5^{00.57}\) & \(60.3^{00.41}\) & \(48.9^{00.17}\) \\
\(k=6\) & \(30.8^{01.76}\) & \(44.9^{00.35}\) & \(60.7^{00.31}\) & \(49.0^{00.17}\) \\
\(k=7\) & \(32.4^{01.85}\) & \(44.8^{00.34}\) & \(60.6^{00.30}\) & \(49.2^{00.12}\) \\
\(k=8\) & \(31.5^{01.52}\) & \(45.0^{00.27}\) & \(60.7^{00.21}\) & \(49.2^{00.13}\) \\
\(k=9\) & \(32.9^{01.41}\) & \(44.8^{00.25}\) & \(60.6^{00.22}\) & \(49.3^{00.10}\) \\
\(k=10\) & \(31.9^{02.16}\) & \(45.0^{00.42}\) & \(60.7^{00.33}\) & \(49.3^{00.08}\) \\
\bottomrule\noalign{}
\end{tabular}
\caption{\label{tbl:rhos_split_top1_accs_vs_k_imagenetlt_crt_cosine}Per-split test top‑1 accuracies for \(\alpha\)‑cRT on ImageNet‑LT using \(k\) nearest neighbors based on cosine distance.}
\end{table}

\hypertarget{tbl:rhos_split_top5_accs_vs_k_imagenetlt_crt_cosine}{}
\begin{table}
\centering
\begin{tabular}[]{@{}lrrrr@{}}

\toprule\noalign{}
Model & Few & Med. & Many & Overall \\
\midrule\noalign{}

cRT & \(57.3\) & \(73.4\) & \(81.8\) & \(74.4\) \\
& & & & \\
\(\bm{\alpha}\)\textbf{‑cRT} & & & & \\
\(\rho=0.25\) & & & & \\
\(k=1\) & \(26.3^{38.14}\) & \(68.0^{10.37}\) & \(78.0^{7.46}\) & \(66.2^{2.65}\) \\
\(k=2\) & \(68.8^{07.12}\) & \(67.5^{06.28}\) & \(77.6^{4.46}\) & \(71.6^{3.76}\) \\
\(k=3\) & \(69.0^{05.03}\) & \(68.7^{04.72}\) & \(78.4^{3.21}\) & \(72.5^{2.84}\) \\
\(k=4\) & \(67.4^{01.50}\) & \(70.5^{00.29}\) & \(79.8^{0.20}\) & \(73.7^{0.20}\) \\
\(k=5\) & \(68.0^{01.83}\) & \(70.8^{00.39}\) & \(79.9^{0.26}\) & \(73.9^{0.14}\) \\
\(k=6\) & \(69.1^{01.09}\) & \(70.6^{00.37}\) & \(79.7^{0.24}\) & \(73.9^{0.15}\) \\
\(k=7\) & \(68.7^{01.30}\) & \(70.7^{00.43}\) & \(79.8^{0.31}\) & \(74.0^{0.18}\) \\
\(k=8\) & \(68.9^{01.45}\) & \(70.7^{00.32}\) & \(79.8^{0.25}\) & \(73.9^{0.09}\) \\
\(k=9\) & \(68.7^{00.82}\) & \(70.8^{00.35}\) & \(79.9^{0.27}\) & \(74.0^{0.18}\) \\
\(k=10\) & \(69.0^{00.69}\) & \(70.6^{00.30}\) & \(79.7^{0.19}\) & \(73.9^{0.21}\) \\
& & & & \\
\(\rho=0.5\) & & & & \\
\(k=1\) & \(42.1^{41.61}\) & \(63.7^{11.32}\) & \(74.9^{8.13}\) & \(65.1^{2.89}\) \\
\(k=2\) & \(64.5^{06.59}\) & \(69.5^{05.12}\) & \(79.0^{3.60}\) & \(72.5^{2.96}\) \\
\(k=3\) & \(64.2^{01.95}\) & \(71.2^{00.48}\) & \(80.1^{0.33}\) & \(73.7^{0.17}\) \\
\(k=4\) & \(65.5^{01.10}\) & \(71.1^{00.24}\) & \(80.1^{0.15}\) & \(73.8^{0.18}\) \\
\(k=5\) & \(65.5^{01.55}\) & \(71.3^{00.31}\) & \(80.2^{0.21}\) & \(73.9^{0.13}\) \\
\(k=6\) & \(65.4^{00.95}\) & \(71.7^{00.26}\) & \(80.5^{0.17}\) & \(74.2^{0.11}\) \\
\(k=7\) & \(66.4^{00.82}\) & \(71.5^{00.25}\) & \(80.3^{0.16}\) & \(74.2^{0.13}\) \\
\(k=8\) & \(66.8^{00.88}\) & \(71.4^{00.28}\) & \(80.2^{0.20}\) & \(74.2^{0.17}\) \\
\(k=9\) & \(66.3^{00.93}\) & \(71.6^{00.28}\) & \(80.4^{0.19}\) & \(74.3^{0.13}\) \\
\(k=10\) & \(66.4^{00.68}\) & \(71.6^{00.19}\) & \(80.4^{0.14}\) & \(74.3^{0.12}\) \\
& & & & \\
\(\rho=1\) & & & & \\
\(k=1\) & \(57.9^{38.14}\) & \(59.4^{10.37}\) & \(71.8^{7.46}\) & \(64.0^{2.65}\) \\
\(k=2\) & \(63.2^{09.28}\) & \(68.9^{06.35}\) & \(78.6^{4.43}\) & \(71.9^{3.50}\) \\
\(k=3\) & \(60.6^{01.44}\) & \(71.9^{00.34}\) & \(80.6^{0.26}\) & \(73.7^{0.12}\) \\
\(k=4\) & \(62.2^{01.02}\) & \(72.0^{00.32}\) & \(80.7^{0.15}\) & \(74.0^{0.17}\) \\
\(k=5\) & \(62.4^{01.48}\) & \(71.8^{00.25}\) & \(80.7^{0.16}\) & \(74.0^{0.12}\) \\
\(k=6\) & \(62.8^{01.10}\) & \(72.2^{00.20}\) & \(80.8^{0.14}\) & \(74.2^{0.09}\) \\
\(k=7\) & \(64.1^{01.35}\) & \(71.9^{00.22}\) & \(80.6^{0.15}\) & \(74.2^{0.11}\) \\
\(k=8\) & \(63.5^{01.03}\) & \(72.1^{00.20}\) & \(80.8^{0.14}\) & \(74.3^{0.09}\) \\
\(k=9\) & \(63.4^{01.06}\) & \(72.1^{00.23}\) & \(80.8^{0.12}\) & \(74.3^{0.12}\) \\
\(k=10\) & \(63.3^{01.22}\) & \(72.2^{00.26}\) & \(80.9^{0.16}\) & \(74.3^{0.07}\) \\
& & & & \\
\(\rho=2\) & & & & \\
\(k=1\) & \(34.2^{40.77}\) & \(65.9^{11.09}\) & \(76.4^{7.97}\) & \(65.6^{2.83}\) \\
\(k=2\) & \(55.7^{01.40}\) & \(72.4^{00.13}\) & \(81.0^{0.09}\) & \(73.4^{0.18}\) \\
\(k=3\) & \(58.1^{01.83}\) & \(72.3^{00.36}\) & \(81.0^{0.23}\) & \(73.7^{0.12}\) \\
\(k=4\) & \(59.4^{01.52}\) & \(72.4^{00.26}\) & \(81.0^{0.19}\) & \(73.9^{0.17}\) \\
\(k=5\) & \(60.2^{01.70}\) & \(72.5^{00.25}\) & \(81.1^{0.19}\) & \(74.2^{0.12}\) \\
\(k=6\) & \(59.3^{01.52}\) & \(72.6^{00.31}\) & \(81.2^{0.16}\) & \(74.1^{0.15}\) \\
\(k=7\) & \(60.5^{01.50}\) & \(72.6^{00.18}\) & \(81.2^{0.15}\) & \(74.3^{0.10}\) \\
\(k=8\) & \(60.0^{01.24}\) & \(72.6^{00.19}\) & \(81.2^{0.11}\) & \(74.2^{0.16}\) \\
\(k=9\) & \(61.0^{00.90}\) & \(72.6^{00.14}\) & \(81.1^{0.10}\) & \(74.3^{0.09}\) \\
\(k=10\) & \(60.5^{01.52}\) & \(72.7^{00.28}\) & \(81.2^{0.16}\) & \(74.3^{0.04}\) \\
\bottomrule\noalign{}
\end{tabular}
\caption{\label{tbl:rhos_split_top5_accs_vs_k_imagenetlt_crt_cosine}Per-split test top‑5 accuracies for \(\alpha\)‑cRT on ImageNet‑LT using \(k\) nearest neighbors based on cosine distance.}
\end{table}

\clearpage

%% file: src/appendix_rhosweep.tex
\hypertarget{sec:rhosweep}{%
\section{Analysis of training data sampling}\label{sec:rhosweep}}

This section contains results for AlphaNet training with a range of \(\rho\) values. Training was performed following the same procedure as described in Section~\ref{sec:impl}. We also include results for the iNaturalist dataset using the cRT model. For iNaturalist, we used smaller values of \(\rho\) given the much smaller differences in per-split accuracy.

\begin{figure}
\hypertarget{fig:models_split_top1_deltas_vs_rho}{%
\centering
\includegraphics{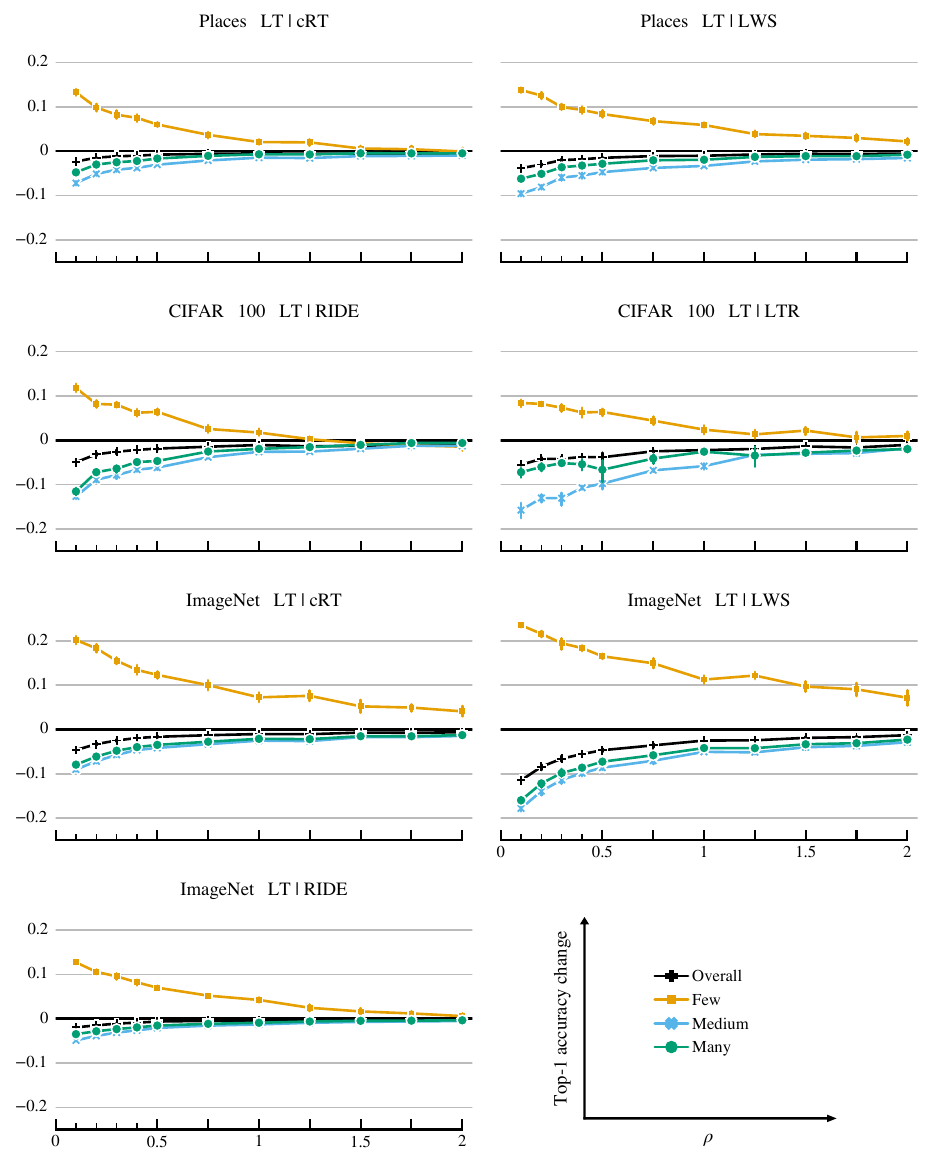}
\caption{Change in per-split top‑1 accuracy vs.~\(\rho\) for AlphaNet training with different models and datasets.}\label{fig:models_split_top1_deltas_vs_rho}
}
\end{figure}

\begin{figure}
\hypertarget{fig:models_split_top5_deltas_vs_rho}{%
\centering
\includegraphics{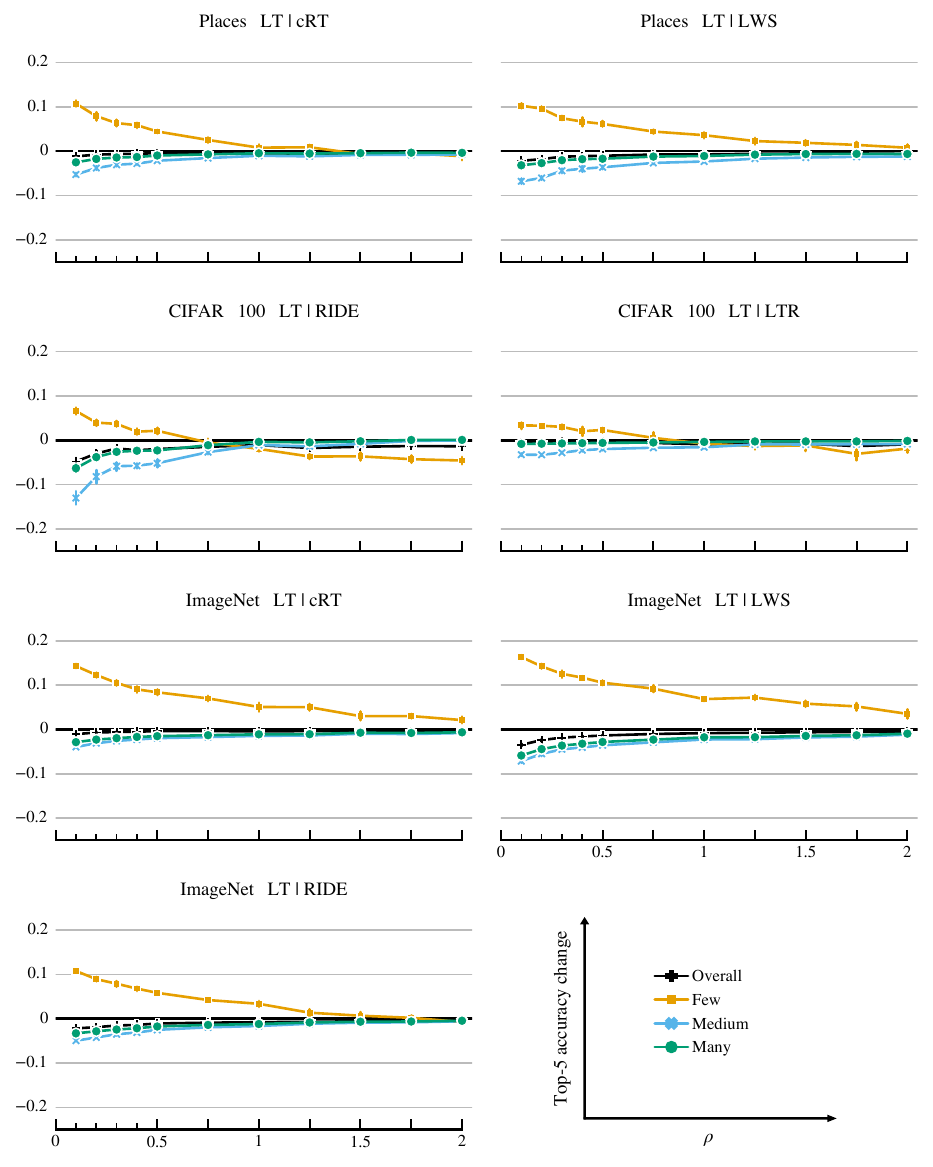}
\caption{Change in per-split top‑5 accuracy vs.~\(\rho\) for AlphaNet training with different models and datasets.}\label{fig:models_split_top5_deltas_vs_rho}
}
\end{figure}

The results are summarized in Figure~\ref{fig:models_split_top1_deltas_vs_rho}, \ref{fig:models_split_top5_deltas_vs_rho}, which show change in per-split top‑1 and top‑5 accuracy respectively, versus \(\rho\) (iNaturalist results are omitted from these figures due to the different set of \(\rho\)s used).

Detailed results, organized by dataset, are shown in the following tables:

\begin{itemize}
\tightlist
\item
  ImageNet‑LT:

  \begin{itemize}
  \tightlist
  \item
    Top‑1 accuracy: Table~\ref{tbl:models_split_top1_accs_vs_rho_imagenetlt}.
  \item
    Top‑5 accuracy: Table~\ref{tbl:models_split_top5_accs_vs_rho_imagenetlt}.
  \end{itemize}
\item
  Places‑LT:

  \begin{itemize}
  \tightlist
  \item
    Top‑1 accuracy: Table~\ref{tbl:models_split_top1_accs_vs_rho_placeslt}.
  \item
    Top‑5 accuracy: Table~\ref{tbl:models_split_top5_accs_vs_rho_placeslt}.
  \end{itemize}
\item
  CIFAR‑100‑LT:

  \begin{itemize}
  \tightlist
  \item
    Top‑1 accuracy: Table~\ref{tbl:models_split_top1_accs_vs_rho_cifarlt}.
  \item
    Top‑5 accuracy: Table~\ref{tbl:models_split_top5_accs_vs_rho_cifarlt}.
  \end{itemize}
\item
  iNaturalist:

  \begin{itemize}
  \tightlist
  \item
    Top‑1 accuracy: Table~\ref{tbl:models_split_top1_accs_vs_rho_inat}.
  \item
    Top‑5 accuracy: Table~\ref{tbl:models_split_top5_accs_vs_rho_inat}.
  \end{itemize}
\end{itemize}

In addition to top‑1 and top‑5 accuracy, we evaluated performance on ImageNet‑LT by considering predictions to a WordNet\citep{2010.Princeton} nearest neighbor as correct.\footnote{This is only possible for ImageNet‑LT since image labels correspond to WordNet synsets.} Given a level \(l\), if the predicted class for a sample is within \(l\) nodes of the true class in the WordNet hierarchy (using the shortest path), it is considered correct. We used \(l=4\), and these results are shown in Table~\ref{tbl:models_split_semantic4_accs_vs_rho_imagenetlt}.

\clearpage

\clearpage

\clearpage

\hypertarget{tbl:models_split_top1_accs_vs_rho_imagenetlt}{}
\begin{table}
\centering
\begin{tabular}[]{@{}lrrrr@{}}

\toprule\noalign{}
Model & Few & Med. & Many & Overall \\
\midrule\noalign{}

cRT & \(27.4\) & \(46.2\) & \(61.8\) & \(49.6\) \\
\(\bm{\alpha}\)\textbf{‑cRT} & & & & \\
\(\rho=0.1\) & \(47.6^{1.60}\) & \(37.1^{0.76}\) & \(53.8^{0.62}\) & \(45.0^{0.41}\) \\
\(\rho=0.2\) & \(45.7^{1.56}\) & \(39.0^{0.78}\) & \(55.7^{0.79}\) & \(46.3^{0.52}\) \\
\(\rho=0.3\) & \(42.9^{1.29}\) & \(40.4^{0.72}\) & \(57.0^{0.68}\) & \(47.1^{0.47}\) \\
\(\rho=0.4\) & \(40.8^{1.85}\) & \(41.5^{0.72}\) & \(57.8^{0.54}\) & \(47.7^{0.36}\) \\
\(\rho=0.5\) & \(39.7^{1.42}\) & \(42.0^{0.66}\) & \(58.3^{0.52}\) & \(48.0^{0.37}\) \\
\(\rho=0.75\) & \(37.4^{1.93}\) & \(42.9^{0.46}\) & \(59.0^{0.40}\) & \(48.3^{0.16}\) \\
\(\rho=1\) & \(34.6^{1.88}\) & \(43.7^{0.51}\) & \(59.7^{0.43}\) & \(48.6^{0.24}\) \\
\(\rho=1.25\) & \(35.0^{1.98}\) & \(43.6^{0.70}\) & \(59.6^{0.50}\) & \(48.6^{0.35}\) \\
\(\rho=1.5\) & \(32.6^{2.46}\) & \(44.4^{0.49}\) & \(60.3^{0.38}\) & \(48.9^{0.19}\) \\
\(\rho=1.75\) & \(32.3^{1.42}\) & \(44.4^{0.32}\) & \(60.3^{0.18}\) & \(48.9^{0.14}\) \\
\(\rho=2\) & \(31.5^{1.99}\) & \(44.7^{0.46}\) & \(60.5^{0.30}\) & \(49.0^{0.12}\) \\
\(\rho=3\) & \(29.0^{2.05}\) & \(45.1^{0.36}\) & \(60.9^{0.28}\) & \(49.0^{0.08}\) \\
& & & & \\
LWS & \(30.4\) & \(47.2\) & \(60.2\) & \(49.9\) \\
\(\bm{\alpha}\)\textbf{‑LWS} & & & & \\
\(\rho=0.1\) & \(53.9^{0.77}\) & \(29.4^{1.22}\) & \(44.2^{1.22}\) & \(38.5^{1.03}\) \\
\(\rho=0.2\) & \(52.0^{1.21}\) & \(33.3^{1.44}\) & \(48.0^{1.37}\) & \(41.5^{1.08}\) \\
\(\rho=0.3\) & \(49.8^{2.20}\) & \(35.9^{1.59}\) & \(50.4^{1.47}\) & \(43.4^{1.10}\) \\
\(\rho=0.4\) & \(48.7^{1.17}\) & \(37.4^{1.13}\) & \(51.6^{0.95}\) & \(44.4^{0.80}\) \\
\(\rho=0.5\) & \(46.9^{0.98}\) & \(38.6^{0.87}\) & \(52.9^{0.86}\) & \(45.3^{0.69}\) \\
\(\rho=0.75\) & \(45.3^{1.89}\) & \(40.2^{1.28}\) & \(54.4^{1.01}\) & \(46.3^{0.76}\) \\
\(\rho=1\) & \(41.6^{1.61}\) & \(42.2^{0.53}\) & \(56.0^{0.32}\) & \(47.4^{0.30}\) \\
\(\rho=1.25\) & \(42.5^{1.41}\) & \(42.1^{0.74}\) & \(56.0^{0.53}\) & \(47.5^{0.41}\) \\
\(\rho=1.5\) & \(40.1^{1.99}\) & \(43.2^{0.98}\) & \(56.9^{0.76}\) & \(48.0^{0.53}\) \\
\(\rho=1.75\) & \(39.4^{2.53}\) & \(43.5^{0.88}\) & \(57.1^{0.70}\) & \(48.2^{0.45}\) \\
\(\rho=2\) & \(37.5^{2.94}\) & \(44.3^{0.86}\) & \(57.9^{0.72}\) & \(48.6^{0.32}\) \\
\(\rho=3\) & \(34.5^{1.91}\) & \(45.3^{0.54}\) & \(58.7^{0.37}\) & \(49.0^{0.21}\) \\
& & & & \\
RIDE & \(36.5\) & \(54.4\) & \(68.9\) & \(57.5\) \\
\(\bm{\alpha}\)\textbf{‑RIDE} & & & & \\
\(\rho=0.1\) & \(49.2^{0.69}\) & \(49.5^{0.40}\) & \(65.4^{0.16}\) & \(55.6^{0.19}\) \\
\(\rho=0.2\) & \(47.1^{0.86}\) & \(50.5^{0.34}\) & \(66.0^{0.20}\) & \(56.0^{0.16}\) \\
\(\rho=0.3\) & \(46.1^{1.16}\) & \(51.2^{0.50}\) & \(66.5^{0.29}\) & \(56.4^{0.21}\) \\
\(\rho=0.4\) & \(44.7^{1.13}\) & \(51.7^{0.39}\) & \(66.9^{0.24}\) & \(56.6^{0.22}\) \\
\(\rho=0.5\) & \(43.5^{0.75}\) & \(52.3^{0.26}\) & \(67.3^{0.17}\) & \(56.9^{0.11}\) \\
\(\rho=0.75\) & \(41.7^{0.63}\) & \(52.8^{0.18}\) & \(67.7^{0.15}\) & \(57.0^{0.12}\) \\
\(\rho=1\) & \(40.8^{1.00}\) & \(53.1^{0.21}\) & \(67.9^{0.18}\) & \(57.1^{0.11}\) \\
\(\rho=1.25\) & \(39.0^{1.28}\) & \(53.4^{0.22}\) & \(68.2^{0.16}\) & \(57.2^{0.05}\) \\
\(\rho=1.5\) & \(38.2^{1.22}\) & \(53.6^{0.25}\) & \(68.4^{0.17}\) & \(57.2^{0.06}\) \\
\(\rho=1.75\) & \(37.7^{0.89}\) & \(53.7^{0.14}\) & \(68.4^{0.10}\) & \(57.2^{0.05}\) \\
\(\rho=2\) & \(37.1^{0.98}\) & \(53.8^{0.12}\) & \(68.5^{0.11}\) & \(57.2^{0.06}\) \\
\(\rho=3\) & \(34.5^{1.23}\) & \(54.3^{0.14}\) & \(68.8^{0.11}\) & \(57.2^{0.08}\) \\
\bottomrule\noalign{}
\end{tabular}
\caption{\label{tbl:models_split_top1_accs_vs_rho_imagenetlt}Top‑1 accuracy on ImageNet‑LT, using AlphaNet applied to different models.}
\end{table}

\hypertarget{tbl:models_split_top5_accs_vs_rho_imagenetlt}{}
\begin{table}
\centering
\begin{tabular}[]{@{}lrrrr@{}}

\toprule\noalign{}
Model & Few & Med. & Many & Overall \\
\midrule\noalign{}

cRT & \(57.3\) & \(73.4\) & \(81.8\) & \(74.4\) \\
\(\bm{\alpha}\)\textbf{‑cRT} & & & & \\
\(\rho=0.1\) & \(71.7^{0.83}\) & \(69.4^{0.25}\) & \(78.9^{0.16}\) & \(73.4^{0.11}\) \\
\(\rho=0.2\) & \(69.6^{0.93}\) & \(70.3^{0.41}\) & \(79.5^{0.27}\) & \(73.7^{0.21}\) \\
\(\rho=0.3\) & \(67.9^{1.03}\) & \(70.8^{0.45}\) & \(79.8^{0.31}\) & \(73.9^{0.21}\) \\
\(\rho=0.4\) & \(66.4^{1.27}\) & \(71.1^{0.36}\) & \(80.1^{0.23}\) & \(73.9^{0.19}\) \\
\(\rho=0.5\) & \(65.7^{1.08}\) & \(71.4^{0.39}\) & \(80.3^{0.25}\) & \(74.0^{0.16}\) \\
\(\rho=0.75\) & \(64.4^{1.08}\) & \(71.6^{0.21}\) & \(80.5^{0.14}\) & \(74.1^{0.10}\) \\
\(\rho=1\) & \(62.4^{1.71}\) & \(71.9^{0.29}\) & \(80.7^{0.20}\) & \(74.0^{0.16}\) \\
\(\rho=1.25\) & \(62.4^{1.26}\) & \(72.0^{0.37}\) & \(80.7^{0.26}\) & \(74.1^{0.18}\) \\
\(\rho=1.5\) & \(60.4^{1.67}\) & \(72.4^{0.24}\) & \(81.0^{0.18}\) & \(74.1^{0.17}\) \\
\(\rho=1.75\) & \(60.4^{0.81}\) & \(72.3^{0.19}\) & \(81.0^{0.11}\) & \(74.0^{0.13}\) \\
\(\rho=2\) & \(59.5^{1.30}\) & \(72.6^{0.25}\) & \(81.1^{0.19}\) & \(74.1^{0.14}\) \\
\(\rho=3\) & \(57.4^{1.45}\) & \(72.9^{0.21}\) & \(81.4^{0.17}\) & \(74.0^{0.11}\) \\
& & & & \\
LWS & \(61.5\) & \(73.7\) & \(81.6\) & \(75.1\) \\
\(\bm{\alpha}\)\textbf{‑LWS} & & & & \\
\(\rho=0.1\) & \(77.9^{0.68}\) & \(66.5^{0.59}\) & \(75.7^{0.65}\) & \(71.6^{0.50}\) \\
\(\rho=0.2\) & \(75.7^{0.90}\) & \(68.2^{0.73}\) & \(77.1^{0.67}\) & \(72.7^{0.50}\) \\
\(\rho=0.3\) & \(74.1^{1.46}\) & \(69.2^{0.67}\) & \(77.9^{0.60}\) & \(73.2^{0.42}\) \\
\(\rho=0.4\) & \(73.2^{0.62}\) & \(69.6^{0.49}\) & \(78.3^{0.39}\) & \(73.5^{0.35}\) \\
\(\rho=0.5\) & \(72.1^{0.90}\) & \(70.1^{0.49}\) & \(78.7^{0.46}\) & \(73.7^{0.32}\) \\
\(\rho=0.75\) & \(70.7^{1.35}\) & \(70.8^{0.48}\) & \(79.3^{0.41}\) & \(74.0^{0.22}\) \\
\(\rho=1\) & \(68.4^{0.76}\) & \(71.4^{0.34}\) & \(79.8^{0.23}\) & \(74.2^{0.21}\) \\
\(\rho=1.25\) & \(68.7^{1.04}\) & \(71.5^{0.35}\) & \(79.8^{0.30}\) & \(74.3^{0.19}\) \\
\(\rho=1.5\) & \(67.3^{1.14}\) & \(71.9^{0.45}\) & \(80.1^{0.36}\) & \(74.4^{0.23}\) \\
\(\rho=1.75\) & \(66.7^{1.48}\) & \(72.1^{0.38}\) & \(80.3^{0.29}\) & \(74.5^{0.21}\) \\
\(\rho=2\) & \(65.0^{1.78}\) & \(72.5^{0.32}\) & \(80.6^{0.24}\) & \(74.6^{0.15}\) \\
\(\rho=3\) & \(63.2^{0.78}\) & \(72.8^{0.23}\) & \(80.8^{0.15}\) & \(74.6^{0.18}\) \\
& & & & \\
RIDE & \(67.9\) & \(79.4\) & \(85.0\) & \(80.0\) \\
\(\bm{\alpha}\)\textbf{‑RIDE} & & & & \\
\(\rho=0.1\) & \(78.6^{0.48}\) & \(74.4^{0.26}\) & \(81.7^{0.16}\) & \(77.8^{0.13}\) \\
\(\rho=0.2\) & \(76.8^{0.82}\) & \(75.1^{0.27}\) & \(82.2^{0.22}\) & \(78.1^{0.14}\) \\
\(\rho=0.3\) & \(75.8^{1.11}\) & \(75.9^{0.52}\) & \(82.6^{0.33}\) & \(78.4^{0.23}\) \\
\(\rho=0.4\) & \(74.7^{0.89}\) & \(76.3^{0.35}\) & \(82.9^{0.24}\) & \(78.6^{0.20}\) \\
\(\rho=0.5\) & \(73.7^{0.54}\) & \(76.9^{0.10}\) & \(83.3^{0.07}\) & \(78.9^{0.09}\) \\
\(\rho=0.75\) & \(72.1^{0.56}\) & \(77.4^{0.25}\) & \(83.6^{0.18}\) & \(79.1^{0.14}\) \\
\(\rho=1\) & \(71.2^{1.16}\) & \(77.7^{0.30}\) & \(83.8^{0.24}\) & \(79.2^{0.17}\) \\
\(\rho=1.25\) & \(69.2^{1.35}\) & \(78.2^{0.23}\) & \(84.2^{0.13}\) & \(79.3^{0.12}\) \\
\(\rho=1.5\) & \(68.6^{1.41}\) & \(78.5^{0.24}\) & \(84.4^{0.17}\) & \(79.4^{0.11}\) \\
\(\rho=1.75\) & \(68.1^{0.99}\) & \(78.6^{0.15}\) & \(84.4^{0.11}\) & \(79.4^{0.10}\) \\
\(\rho=2\) & \(67.1^{1.05}\) & \(78.8^{0.16}\) & \(84.6^{0.12}\) & \(79.4^{0.08}\) \\
\(\rho=3\) & \(63.7^{1.45}\) & \(79.3^{0.18}\) & \(84.9^{0.09}\) & \(79.3^{0.10}\) \\
\bottomrule\noalign{}
\end{tabular}
\caption{\label{tbl:models_split_top5_accs_vs_rho_imagenetlt}Top‑5 accuracy on ImageNet‑LT, using AlphaNet applied to different models.}
\end{table}

\hypertarget{tbl:models_split_semantic4_accs_vs_rho_imagenetlt}{}
\begin{table}
\centering
\begin{tabular}[]{@{}lrrrr@{}}

\toprule\noalign{}
Model & Few & Med. & Many & Overall \\
\midrule\noalign{}

cRT & \(46.5\) & \(59.7\) & \(71.0\) & \(62.3\) \\
\(\bm{\alpha}\)\textbf{‑cRT} & & & & \\
\(\rho=0.1\) & \(57.8^{1.02}\) & \(53.5^{0.48}\) & \(65.7^{0.39}\) & \(58.8^{0.27}\) \\
\(\rho=0.2\) & \(56.6^{0.82}\) & \(54.9^{0.51}\) & \(66.7^{0.53}\) & \(59.7^{0.38}\) \\
\(\rho=0.3\) & \(55.0^{0.74}\) & \(55.8^{0.65}\) & \(67.8^{0.40}\) & \(60.3^{0.39}\) \\
\(\rho=0.4\) & \(53.9^{1.02}\) & \(56.5^{0.55}\) & \(68.3^{0.40}\) & \(60.7^{0.31}\) \\
\(\rho=0.5\) & \(53.2^{0.72}\) & \(56.9^{0.49}\) & \(68.7^{0.32}\) & \(60.9^{0.29}\) \\
\(\rho=0.75\) & \(51.8^{1.21}\) & \(57.3^{0.35}\) & \(69.1^{0.29}\) & \(61.1^{0.15}\) \\
\(\rho=1\) & \(50.4^{1.06}\) & \(57.7^{0.35}\) & \(69.6^{0.27}\) & \(61.3^{0.18}\) \\
\(\rho=1.25\) & \(50.6^{1.09}\) & \(57.8^{0.49}\) & \(69.6^{0.34}\) & \(61.3^{0.27}\) \\
\(\rho=1.5\) & \(49.2^{1.39}\) & \(58.4^{0.28}\) & \(70.0^{0.25}\) & \(61.6^{0.13}\) \\
\(\rho=1.75\) & \(49.0^{0.81}\) & \(58.4^{0.17}\) & \(70.0^{0.13}\) & \(61.6^{0.11}\) \\
\(\rho=2\) & \(48.5^{1.06}\) & \(58.7^{0.33}\) & \(70.1^{0.18}\) & \(61.7^{0.10}\) \\
\(\rho=3\) & \(47.0^{1.18}\) & \(58.9^{0.23}\) & \(70.4^{0.18}\) & \(61.7^{0.06}\) \\
& & & & \\
LWS & \(48.3\) & \(60.4\) & \(69.8\) & \(62.4\) \\
\(\bm{\alpha}\)\textbf{‑LWS} & & & & \\
\(\rho=0.1\) & \(61.5^{0.55}\) & \(47.7^{1.08}\) & \(58.9^{0.62}\) & \(53.9^{0.74}\) \\
\(\rho=0.2\) & \(60.4^{0.94}\) & \(50.6^{1.18}\) & \(61.5^{0.89}\) & \(56.1^{0.80}\) \\
\(\rho=0.3\) & \(59.1^{1.18}\) & \(52.5^{1.05}\) & \(63.2^{0.99}\) & \(57.5^{0.74}\) \\
\(\rho=0.4\) & \(58.4^{0.58}\) & \(53.7^{0.73}\) & \(64.1^{0.60}\) & \(58.3^{0.52}\) \\
\(\rho=0.5\) & \(57.5^{0.60}\) & \(54.4^{0.91}\) & \(64.8^{0.62}\) & \(58.8^{0.64}\) \\
\(\rho=0.75\) & \(56.5^{1.06}\) & \(55.6^{1.00}\) & \(65.9^{0.66}\) & \(59.7^{0.59}\) \\
\(\rho=1\) & \(54.4^{0.89}\) & \(56.9^{0.36}\) & \(67.0^{0.18}\) & \(60.5^{0.23}\) \\
\(\rho=1.25\) & \(55.0^{0.85}\) & \(56.8^{0.59}\) & \(67.0^{0.38}\) & \(60.5^{0.33}\) \\
\(\rho=1.5\) & \(53.6^{1.16}\) & \(57.8^{0.76}\) & \(67.5^{0.49}\) & \(60.9^{0.41}\) \\
\(\rho=1.75\) & \(53.1^{1.31}\) & \(57.9^{0.59}\) & \(67.8^{0.45}\) & \(61.1^{0.32}\) \\
\(\rho=2\) & \(52.1^{1.36}\) & \(58.5^{0.45}\) & \(68.3^{0.48}\) & \(61.4^{0.24}\) \\
\(\rho=3\) & \(50.7^{0.80}\) & \(59.1^{0.29}\) & \(68.8^{0.23}\) & \(61.7^{0.15}\) \\
& & & & \\
RIDE & \(53.8\) & \(66.4\) & \(76.3\) & \(68.5\) \\
\(\bm{\alpha}\)\textbf{‑RIDE} & & & & \\
\(\rho=0.1\) & \(61.6^{0.46}\) & \(62.9^{0.30}\) & \(73.7^{0.11}\) & \(66.9^{0.17}\) \\
\(\rho=0.2\) & \(60.4^{0.48}\) & \(63.7^{0.24}\) & \(74.3^{0.19}\) & \(67.3^{0.14}\) \\
\(\rho=0.3\) & \(59.7^{0.76}\) & \(64.2^{0.40}\) & \(74.6^{0.27}\) & \(67.6^{0.21}\) \\
\(\rho=0.4\) & \(58.9^{0.70}\) & \(64.5^{0.28}\) & \(74.9^{0.22}\) & \(67.8^{0.20}\) \\
\(\rho=0.5\) & \(58.2^{0.39}\) & \(65.0^{0.13}\) & \(75.2^{0.09}\) & \(68.0^{0.08}\) \\
\(\rho=0.75\) & \(57.0^{0.39}\) & \(65.4^{0.18}\) & \(75.5^{0.13}\) & \(68.1^{0.13}\) \\
\(\rho=1\) & \(56.5^{0.74}\) & \(65.5^{0.17}\) & \(75.7^{0.14}\) & \(68.2^{0.11}\) \\
\(\rho=1.25\) & \(55.3^{0.90}\) & \(65.8^{0.14}\) & \(75.9^{0.12}\) & \(68.2^{0.07}\) \\
\(\rho=1.5\) & \(54.6^{0.88}\) & \(65.9^{0.15}\) & \(76.0^{0.12}\) & \(68.3^{0.04}\) \\
\(\rho=1.75\) & \(54.5^{0.62}\) & \(66.0^{0.08}\) & \(76.0^{0.06}\) & \(68.3^{0.06}\) \\
\(\rho=2\) & \(54.1^{0.70}\) & \(66.0^{0.10}\) & \(76.1^{0.11}\) & \(68.3^{0.04}\) \\
\(\rho=3\) & \(52.2^{0.76}\) & \(66.3^{0.07}\) & \(76.3^{0.09}\) & \(68.2^{0.06}\) \\
\bottomrule\noalign{}
\end{tabular}
\caption{\label{tbl:models_split_semantic4_accs_vs_rho_imagenetlt}ImageNet‑LT accuracy computed by considering predictions within 4 WordNet nodes as correct, for AlphaNet applied to different models.}
\end{table}

\clearpage

\hypertarget{tbl:models_split_top1_accs_vs_rho_placeslt}{}
\begin{table}
\centering
\begin{tabular}[]{@{}lrrrr@{}}

\toprule\noalign{}
Model & Few & Med. & Many & Overall \\
\midrule\noalign{}

cRT & \(24.9\) & \(37.6\) & \(42.0\) & \(36.7\) \\
\(\bm{\alpha}\)\textbf{‑cRT} & & & & \\
\(\rho=0.1\) & \(38.2^{1.30}\) & \(30.4^{0.84}\) & \(37.3^{0.72}\) & \(34.4^{0.41}\) \\
\(\rho=0.2\) & \(34.8^{1.42}\) & \(32.4^{0.66}\) & \(39.0^{0.29}\) & \(35.3^{0.15}\) \\
\(\rho=0.3\) & \(33.2^{1.57}\) & \(33.4^{0.75}\) & \(39.5^{0.60}\) & \(35.6^{0.25}\) \\
\(\rho=0.4\) & \(32.4^{1.47}\) & \(33.8^{0.69}\) & \(39.8^{0.46}\) & \(35.7^{0.22}\) \\
\(\rho=0.5\) & \(31.0^{0.88}\) & \(34.5^{0.17}\) & \(40.4^{0.29}\) & \(35.9^{0.09}\) \\
\(\rho=0.75\) & \(28.6^{1.27}\) & \(35.5^{0.40}\) & \(41.0^{0.13}\) & \(36.1^{0.11}\) \\
\(\rho=1\) & \(27.0^{1.02}\) & \(36.1^{0.31}\) & \(41.3^{0.13}\) & \(36.2^{0.10}\) \\
\(\rho=1.25\) & \(26.9^{1.23}\) & \(36.0^{0.47}\) & \(41.3^{0.14}\) & \(36.1^{0.16}\) \\
\(\rho=1.5\) & \(25.5^{0.89}\) & \(36.5^{0.36}\) & \(41.6^{0.21}\) & \(36.2^{0.11}\) \\
\(\rho=1.75\) & \(25.4^{1.32}\) & \(36.5^{0.29}\) & \(41.5^{0.23}\) & \(36.2^{0.11}\) \\
\(\rho=2\) & \(24.9^{1.38}\) & \(36.6^{0.49}\) & \(41.5^{0.20}\) & \(36.1^{0.10}\) \\
\(\rho=3\) & \(22.3^{1.56}\) & \(37.3^{0.28}\) & \(41.9^{0.20}\) & \(36.1^{0.14}\) \\
& & & & \\
LWS & \(28.7\) & \(39.1\) & \(40.6\) & \(37.6\) \\
\(\bm{\alpha}\)\textbf{‑LWS} & & & & \\
\(\rho=0.1\) & \(42.5^{1.03}\) & \(29.5^{1.01}\) & \(34.4^{0.94}\) & \(33.8^{0.58}\) \\
\(\rho=0.2\) & \(41.3^{1.30}\) & \(31.0^{0.81}\) & \(35.5^{0.81}\) & \(34.6^{0.44}\) \\
\(\rho=0.3\) & \(38.7^{1.13}\) & \(33.2^{1.12}\) & \(37.0^{0.70}\) & \(35.6^{0.51}\) \\
\(\rho=0.4\) & \(38.0^{1.45}\) & \(33.6^{1.18}\) & \(37.3^{0.65}\) & \(35.8^{0.50}\) \\
\(\rho=0.5\) & \(37.1^{1.39}\) & \(34.4^{0.80}\) & \(37.7^{0.52}\) & \(36.1^{0.31}\) \\
\(\rho=0.75\) & \(35.5^{1.25}\) & \(35.3^{0.59}\) & \(38.5^{0.28}\) & \(36.5^{0.16}\) \\
\(\rho=1\) & \(34.6^{0.97}\) & \(35.8^{0.54}\) & \(38.6^{0.39}\) & \(36.6^{0.22}\) \\
\(\rho=1.25\) & \(32.6^{1.28}\) & \(36.8^{0.60}\) & \(39.3^{0.39}\) & \(36.9^{0.19}\) \\
\(\rho=1.5\) & \(32.2^{1.17}\) & \(37.2^{0.36}\) & \(39.5^{0.39}\) & \(37.0^{0.11}\) \\
\(\rho=1.75\) & \(31.7^{1.35}\) & \(37.3^{0.43}\) & \(39.5^{0.27}\) & \(37.0^{0.09}\) \\
\(\rho=2\) & \(30.9^{1.28}\) & \(37.6^{0.40}\) & \(39.8^{0.18}\) & \(37.1^{0.14}\) \\
\(\rho=3\) & \(27.8^{1.94}\) & \(38.5^{0.51}\) & \(40.2^{0.29}\) & \(37.1^{0.10}\) \\
\bottomrule\noalign{}
\end{tabular}
\caption{\label{tbl:models_split_top1_accs_vs_rho_placeslt}Top‑1 accuracy on Places‑LT, using AlphaNet applied to different models.}
\end{table}

\hypertarget{tbl:models_split_top5_accs_vs_rho_placeslt}{}
\begin{table}
\centering
\begin{tabular}[]{@{}lrrrr@{}}

\toprule\noalign{}
Model & Few & Med. & Many & Overall \\
\midrule\noalign{}

cRT & \(56.3\) & \(70.2\) & \(74.0\) & \(68.9\) \\
\(\bm{\alpha}\)\textbf{‑cRT} & & & & \\
\(\rho=0.1\) & \(67.1^{1.22}\) & \(64.9^{0.69}\) & \(71.5^{0.30}\) & \(67.7^{0.32}\) \\
\(\rho=0.2\) & \(64.2^{1.61}\) & \(66.4^{0.74}\) & \(72.3^{0.35}\) & \(68.1^{0.25}\) \\
\(\rho=0.3\) & \(62.7^{1.32}\) & \(67.1^{0.63}\) & \(72.6^{0.34}\) & \(68.2^{0.24}\) \\
\(\rho=0.4\) & \(62.2^{1.06}\) & \(67.4^{0.55}\) & \(72.7^{0.35}\) & \(68.3^{0.23}\) \\
\(\rho=0.5\) & \(60.8^{0.60}\) & \(68.0^{0.19}\) & \(73.0^{0.15}\) & \(68.4^{0.19}\) \\
\(\rho=0.75\) & \(58.9^{1.07}\) & \(68.6^{0.34}\) & \(73.3^{0.20}\) & \(68.4^{0.21}\) \\
\(\rho=1\) & \(57.1^{1.29}\) & \(69.2^{0.30}\) & \(73.5^{0.16}\) & \(68.4^{0.19}\) \\
\(\rho=1.25\) & \(57.2^{0.94}\) & \(69.0^{0.29}\) & \(73.5^{0.19}\) & \(68.3^{0.18}\) \\
\(\rho=1.5\) & \(55.7^{0.85}\) & \(69.4^{0.29}\) & \(73.6^{0.26}\) & \(68.2^{0.22}\) \\
\(\rho=1.75\) & \(55.8^{1.09}\) & \(69.4^{0.25}\) & \(73.7^{0.15}\) & \(68.3^{0.09}\) \\
\(\rho=2\) & \(55.2^{1.49}\) & \(69.5^{0.38}\) & \(73.6^{0.24}\) & \(68.2^{0.20}\) \\
\(\rho=3\) & \(52.4^{1.83}\) & \(70.1^{0.35}\) & \(73.9^{0.15}\) & \(68.0^{0.25}\) \\
& & & & \\
LWS & \(60.2\) & \(70.8\) & \(73.4\) & \(69.7\) \\
\(\bm{\alpha}\)\textbf{‑LWS} & & & & \\
\(\rho=0.1\) & \(70.5^{0.94}\) & \(64.0^{1.02}\) & \(70.2^{0.44}\) & \(67.5^{0.43}\) \\
\(\rho=0.2\) & \(69.8^{0.76}\) & \(64.8^{0.69}\) & \(70.7^{0.39}\) & \(67.9^{0.33}\) \\
\(\rho=0.3\) & \(67.7^{0.86}\) & \(66.4^{0.94}\) & \(71.4^{0.41}\) & \(68.4^{0.43}\) \\
\(\rho=0.4\) & \(66.9^{1.74}\) & \(66.8^{1.15}\) & \(71.6^{0.60}\) & \(68.6^{0.41}\) \\
\(\rho=0.5\) & \(66.4^{1.10}\) & \(67.2^{0.70}\) & \(71.7^{0.41}\) & \(68.6^{0.34}\) \\
\(\rho=0.75\) & \(64.6^{0.81}\) & \(68.2^{0.47}\) & \(72.2^{0.27}\) & \(68.9^{0.20}\) \\
\(\rho=1\) & \(63.8^{1.20}\) & \(68.5^{0.48}\) & \(72.3^{0.34}\) & \(69.0^{0.21}\) \\
\(\rho=1.25\) & \(62.5^{1.31}\) & \(69.1^{0.49}\) & \(72.6^{0.28}\) & \(69.1^{0.13}\) \\
\(\rho=1.5\) & \(62.1^{1.04}\) & \(69.4^{0.25}\) & \(72.8^{0.17}\) & \(69.2^{0.16}\) \\
\(\rho=1.75\) & \(61.6^{1.17}\) & \(69.5^{0.44}\) & \(72.8^{0.24}\) & \(69.2^{0.20}\) \\
\(\rho=2\) & \(61.0^{1.33}\) & \(69.6^{0.44}\) & \(72.8^{0.23}\) & \(69.1^{0.15}\) \\
\(\rho=3\) & \(58.3^{2.15}\) & \(70.3^{0.46}\) & \(73.2^{0.26}\) & \(69.0^{0.23}\) \\
\bottomrule\noalign{}
\end{tabular}
\caption{\label{tbl:models_split_top5_accs_vs_rho_placeslt}Top‑5 accuracy on Places‑LT, using AlphaNet applied to different models.}
\end{table}

\clearpage

\hypertarget{tbl:models_split_top1_accs_vs_rho_cifarlt}{}
\begin{table}
\centering
\begin{tabular}[]{@{}lrrrr@{}}

\toprule\noalign{}
Model & Few & Med. & Many & Overall \\
\midrule\noalign{}

RIDE & \(25.8\) & \(52.1\) & \(69.3\) & \(50.2\) \\
\(\bm{\alpha}\)\textbf{‑RIDE} & & & & \\
\(\rho=0.1\) & \(37.6^{1.64}\) & \(39.4^{0.99}\) & \(57.8^{1.40}\) & \(45.3^{0.40}\) \\
\(\rho=0.2\) & \(34.0^{1.42}\) & \(43.2^{1.01}\) & \(62.1^{0.64}\) & \(47.1^{0.58}\) \\
\(\rho=0.3\) & \(33.9^{1.02}\) & \(44.2^{1.37}\) & \(62.9^{1.40}\) & \(47.7^{0.87}\) \\
\(\rho=0.4\) & \(32.1^{1.47}\) & \(45.5^{0.74}\) & \(64.4^{0.83}\) & \(48.1^{0.28}\) \\
\(\rho=0.5\) & \(32.3^{1.24}\) & \(45.9^{0.87}\) & \(64.6^{0.78}\) & \(48.4^{0.43}\) \\
\(\rho=0.75\) & \(28.4^{1.56}\) & \(48.3^{0.58}\) & \(66.8^{0.37}\) & \(48.8^{0.33}\) \\
\(\rho=1\) & \(27.6^{1.41}\) & \(49.5^{0.83}\) & \(67.4^{0.70}\) & \(49.2^{0.16}\) \\
\(\rho=1.25\) & \(26.1^{0.98}\) & \(49.5^{0.53}\) & \(67.8^{0.21}\) & \(48.9^{0.35}\) \\
\(\rho=1.5\) & \(25.2^{1.11}\) & \(50.2^{0.57}\) & \(68.3^{0.34}\) & \(49.0^{0.26}\) \\
\(\rho=1.75\) & \(24.7^{1.56}\) & \(50.9^{0.62}\) & \(68.7^{0.49}\) & \(49.3^{0.19}\) \\
\(\rho=2\) & \(24.4^{1.43}\) & \(50.8^{0.73}\) & \(68.7^{0.54}\) & \(49.2^{0.23}\) \\
\(\rho=3\) & \(22.1^{1.30}\) & \(51.7^{0.68}\) & \(69.3^{0.44}\) & \(49.0^{0.30}\) \\
& & & & \\
LTR & \(29.8\) & \(49.3\) & \(70.1\) & \(50.7\) \\
\(\bm{\alpha}\)\textbf{‑LTR} & & & & \\
\(\rho=0.1\) & \(38.2^{1.42}\) & \(33.5^{2.91}\) & \(62.9^{1.87}\) & \(45.2^{1.10}\) \\
\(\rho=0.2\) & \(38.1^{0.78}\) & \(36.2^{1.52}\) & \(64.1^{1.53}\) & \(46.5^{0.69}\) \\
\(\rho=0.3\) & \(37.2^{1.44}\) & \(36.2^{2.45}\) & \(65.0^{0.93}\) & \(46.6^{1.07}\) \\
\(\rho=0.4\) & \(36.1^{1.90}\) & \(38.6^{0.80}\) & \(64.7^{2.13}\) & \(47.0^{0.76}\) \\
\(\rho=0.5\) & \(36.2^{1.39}\) & \(39.5^{2.11}\) & \(63.4^{4.18}\) & \(46.9^{1.95}\) \\
\(\rho=0.75\) & \(34.3^{1.59}\) & \(42.5^{0.59}\) & \(66.0^{1.78}\) & \(48.2^{0.43}\) \\
\(\rho=1\) & \(32.2^{1.77}\) & \(43.4^{1.22}\) & \(67.5^{0.95}\) & \(48.5^{0.73}\) \\
\(\rho=1.25\) & \(31.2^{1.70}\) & \(46.0^{0.67}\) & \(66.6^{3.91}\) & \(48.8^{1.11}\) \\
\(\rho=1.5\) & \(32.0^{1.49}\) & \(46.2^{0.86}\) & \(67.3^{1.02}\) & \(49.3^{0.33}\) \\
\(\rho=1.75\) & \(30.5^{2.05}\) & \(46.4^{1.00}\) & \(67.8^{0.75}\) & \(49.1^{0.34}\) \\
\(\rho=2\) & \(30.8^{1.72}\) & \(47.5^{1.00}\) & \(68.1^{1.22}\) & \(49.7^{0.40}\) \\
\(\rho=3\) & \(29.9^{1.72}\) & \(49.0^{0.86}\) & \(68.6^{1.01}\) & \(50.1^{0.20}\) \\
\bottomrule\noalign{}
\end{tabular}
\caption{\label{tbl:models_split_top1_accs_vs_rho_cifarlt}Top‑1 accuracy on CIFAR‑100‑LT, using AlphaNet applied to different models.}
\end{table}

\hypertarget{tbl:models_split_top5_accs_vs_rho_cifarlt}{}
\begin{table}
\centering
\begin{tabular}[]{@{}lrrrr@{}}

\toprule\noalign{}
Model & Few & Med. & Many & Overall \\
\midrule\noalign{}

RIDE & \(68.8\) & \(80.6\) & \(86.3\) & \(79.1\) \\
\(\bm{\alpha}\)\textbf{‑RIDE} & & & & \\
\(\rho=0.1\) & \(75.5^{1.34}\) & \(67.6^{2.59}\) & \(80.0^{1.38}\) & \(74.3^{1.33}\) \\
\(\rho=0.2\) & \(72.8^{1.17}\) & \(72.4^{2.56}\) & \(82.5^{1.15}\) & \(76.1^{1.20}\) \\
\(\rho=0.3\) & \(72.5^{1.11}\) & \(74.8^{1.67}\) & \(83.7^{0.71}\) & \(77.2^{0.63}\) \\
\(\rho=0.4\) & \(70.8^{1.24}\) & \(74.9^{1.03}\) & \(84.0^{0.60}\) & \(76.8^{0.50}\) \\
\(\rho=0.5\) & \(71.0^{1.25}\) & \(75.4^{1.58}\) & \(84.0^{0.72}\) & \(77.1^{0.77}\) \\
\(\rho=0.75\) & \(68.3^{1.40}\) & \(77.9^{0.91}\) & \(85.2^{0.24}\) & \(77.6^{0.57}\) \\
\(\rho=1\) & \(66.9^{1.14}\) & \(79.5^{0.53}\) & \(86.0^{0.29}\) & \(78.0^{0.20}\) \\
\(\rho=1.25\) & \(65.2^{0.96}\) & \(79.3^{0.79}\) & \(85.9^{0.33}\) & \(77.3^{0.55}\) \\
\(\rho=1.5\) & \(65.2^{1.55}\) & \(79.8^{0.32}\) & \(86.1^{0.20}\) & \(77.6^{0.50}\) \\
\(\rho=1.75\) & \(64.6^{1.43}\) & \(80.5^{0.50}\) & \(86.4^{0.26}\) & \(77.8^{0.29}\) \\
\(\rho=2\) & \(64.3^{1.45}\) & \(80.6^{0.59}\) & \(86.4^{0.28}\) & \(77.7^{0.26}\) \\
\(\rho=3\) & \(62.0^{1.24}\) & \(81.1^{0.73}\) & \(86.8^{0.38}\) & \(77.4^{0.38}\) \\
& & & & \\
LTR & \(69.3\) & \(72.0\) & \(80.8\) & \(74.3\) \\
\(\bm{\alpha}\)\textbf{‑LTR} & & & & \\
\(\rho=0.1\) & \(72.7^{1.25}\) & \(68.8^{0.47}\) & \(80.0^{0.12}\) & \(73.9^{0.25}\) \\
\(\rho=0.2\) & \(72.6^{0.51}\) & \(68.7^{0.32}\) & \(80.1^{0.07}\) & \(73.9^{0.22}\) \\
\(\rho=0.3\) & \(72.3^{0.98}\) & \(69.2^{0.39}\) & \(80.1^{0.06}\) & \(74.0^{0.33}\) \\
\(\rho=0.4\) & \(71.4^{1.71}\) & \(69.8^{0.49}\) & \(80.2^{0.08}\) & \(73.9^{0.57}\) \\
\(\rho=0.5\) & \(71.6^{0.88}\) & \(70.0^{0.46}\) & \(80.2^{0.11}\) & \(74.1^{0.29}\) \\
\(\rho=0.75\) & \(69.9^{2.14}\) & \(70.3^{0.46}\) & \(80.3^{0.11}\) & \(73.7^{0.71}\) \\
\(\rho=1\) & \(68.5^{1.79}\) & \(70.4^{0.53}\) & \(80.5^{0.17}\) & \(73.4^{0.56}\) \\
\(\rho=1.25\) & \(68.0^{1.39}\) & \(71.1^{0.45}\) & \(80.6^{0.12}\) & \(73.5^{0.46}\) \\
\(\rho=1.5\) & \(68.1^{2.10}\) & \(71.1^{0.47}\) & \(80.6^{0.19}\) & \(73.5^{0.71}\) \\
\(\rho=1.75\) & \(66.2^{2.59}\) & \(71.1^{0.42}\) & \(80.6^{0.18}\) & \(73.0^{0.70}\) \\
\(\rho=2\) & \(67.4^{1.57}\) & \(71.2^{0.47}\) & \(80.7^{0.11}\) & \(73.4^{0.38}\) \\
\(\rho=3\) & \(67.1^{2.00}\) & \(71.7^{0.57}\) & \(80.8^{0.17}\) & \(73.5^{0.48}\) \\
\bottomrule\noalign{}
\end{tabular}
\caption{\label{tbl:models_split_top5_accs_vs_rho_cifarlt}Top‑5 accuracy on CIFAR‑100‑LT, using AlphaNet applied to different models.}
\end{table}

\clearpage

\hypertarget{tbl:models_split_top1_accs_vs_rho_inat}{}
\begin{table}
\centering
\begin{tabular}[]{@{}lrrrr@{}}

\toprule\noalign{}
Model & Few & Med. & Many & Overall \\
\midrule\noalign{}

cRT & \(69.2\) & \(71.9\) & \(75.7\) & \(71.2\) \\
\(\bm{\alpha}\)\textbf{‑cRT} & & & & \\
\(\rho=0.01\) & \(76.1^{0.60}\) & \(54.9^{2.54}\) & \(65.5^{2.11}\) & \(64.4^{1.43}\) \\
\(\rho=0.02\) & \(74.4^{0.77}\) & \(59.1^{1.56}\) & \(68.6^{1.27}\) & \(66.2^{1.05}\) \\
\(\rho=0.03\) & \(74.2^{0.66}\) & \(61.1^{1.22}\) & \(70.0^{0.85}\) & \(67.2^{0.59}\) \\
\(\rho=0.04\) & \(73.7^{0.69}\) & \(61.6^{1.61}\) & \(70.4^{0.94}\) & \(67.3^{0.86}\) \\
\(\rho=0.05\) & \(73.3^{0.68}\) & \(62.8^{1.14}\) & \(71.0^{0.68}\) & \(67.8^{0.69}\) \\
\bottomrule\noalign{}
\end{tabular}
\caption{\label{tbl:models_split_top1_accs_vs_rho_inat}Top‑1 accuracy on iNaturalist, using AlphaNet applied to cRT.}
\end{table}

\hypertarget{tbl:models_split_top5_accs_vs_rho_inat}{}
\begin{table}
\centering
\begin{tabular}[]{@{}lrrrr@{}}

\toprule\noalign{}
Model & Few & Med. & Many & Overall \\
\midrule\noalign{}

cRT & \(87.7\) & \(88.1\) & \(89.8\) & \(88.1\) \\
\(\bm{\alpha}\)\textbf{‑cRT} & & & & \\
\(\rho=0.01\) & \(90.3^{0.33}\) & \(82.3^{1.33}\) & \(86.3^{0.79}\) & \(85.9^{0.65}\) \\
\(\rho=0.02\) & \(89.5^{0.28}\) & \(83.5^{1.20}\) & \(87.2^{0.81}\) & \(86.3^{0.62}\) \\
\(\rho=0.03\) & \(89.2^{0.27}\) & \(84.4^{0.55}\) & \(87.9^{0.32}\) & \(86.7^{0.28}\) \\
\(\rho=0.04\) & \(89.2^{0.25}\) & \(84.4^{0.79}\) & \(87.9^{0.51}\) & \(86.7^{0.40}\) \\
\(\rho=0.05\) & \(88.7^{0.33}\) & \(84.7^{0.67}\) & \(88.1^{0.41}\) & \(86.6^{0.32}\) \\
\bottomrule\noalign{}
\end{tabular}
\caption{\label{tbl:models_split_top5_accs_vs_rho_inat}Top‑5 accuracy on iNaturalist, using AlphaNet applied to cRT.}
\end{table}

\clearpage

%% file: src/appendix_perclsdels.tex
\graphicspath{{figures/appendix/}}

\hypertarget{sec:perclsdels}{%
\section{Change in per-class accuracies}\label{sec:perclsdels}}

\begin{figure}
\hypertarget{fig:rhos_cls_deltas_imagenetlt_crt}{%
\centering
\includegraphics{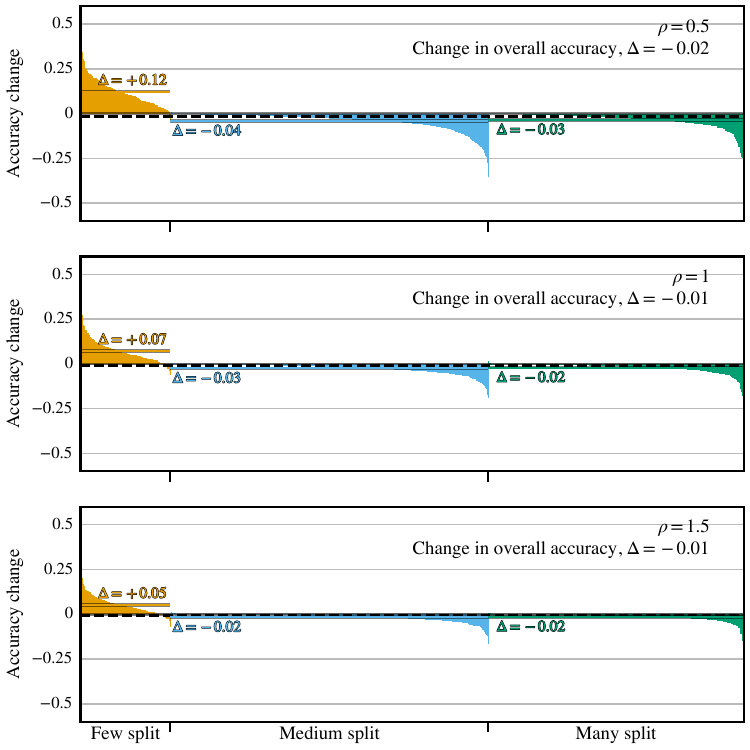}
\caption{Change in per-class test accuracy on ImageNet‑LT after AlphaNet training with cRT baseline. Each bar shows the change the change in accuracy for one class. The solid lines in each split show the average per-class change for the split, and the dotted line shows the overall average per-class change.}\label{fig:rhos_cls_deltas_imagenetlt_crt}
}
\end{figure}

\begin{figure}
\hypertarget{fig:rhos_cls_deltas_imagenetlt_lws}{%
\centering
\includegraphics{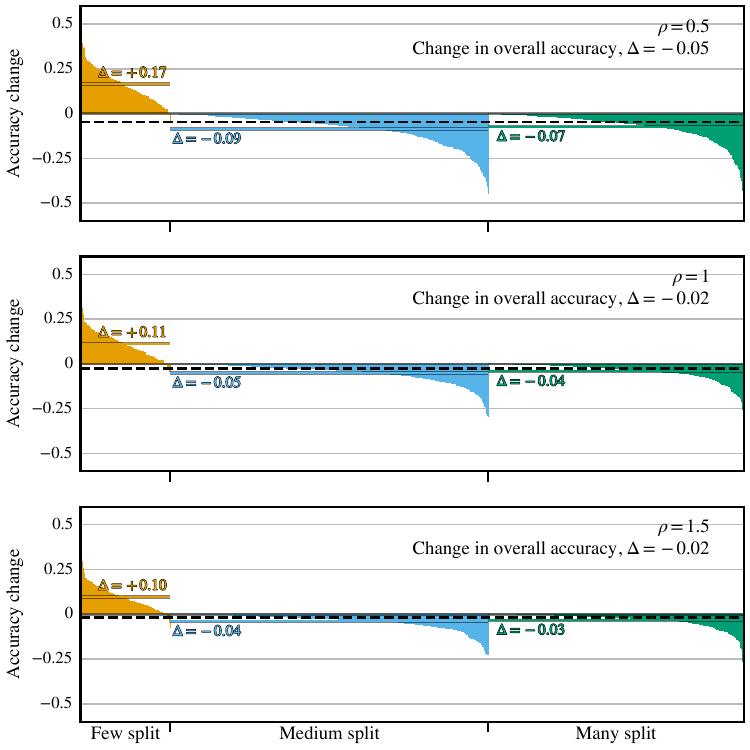}
\caption{Change in per-class test accuracy on ImageNet‑LT after AlphaNet training with LWS baseline. Each bar shows the change the change in accuracy for one class. The solid lines in each split show the average per-class change for the split, and the dotted line shows the overall average per-class change.}\label{fig:rhos_cls_deltas_imagenetlt_lws}
}
\end{figure}

\begin{figure}
\hypertarget{fig:rhos_cls_deltas_imagenetlt_ride}{%
\centering
\includegraphics{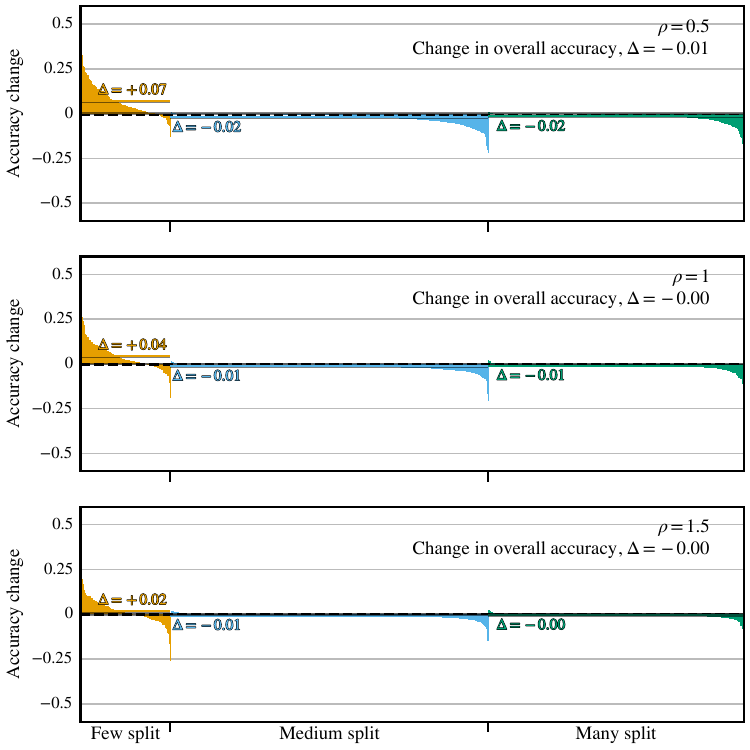}
\caption{Change in per-class test accuracy on ImageNet‑LT after AlphaNet training with RIDE baseline. Each bar shows the change the change in accuracy for one class. The solid lines in each split show the average per-class change for the split, and the dotted line shows the overall average per-class change.}\label{fig:rhos_cls_deltas_imagenetlt_ride}
}
\end{figure}

\begin{figure}
\hypertarget{fig:rhos_cls_deltas_placeslt_crt}{%
\centering
\includegraphics{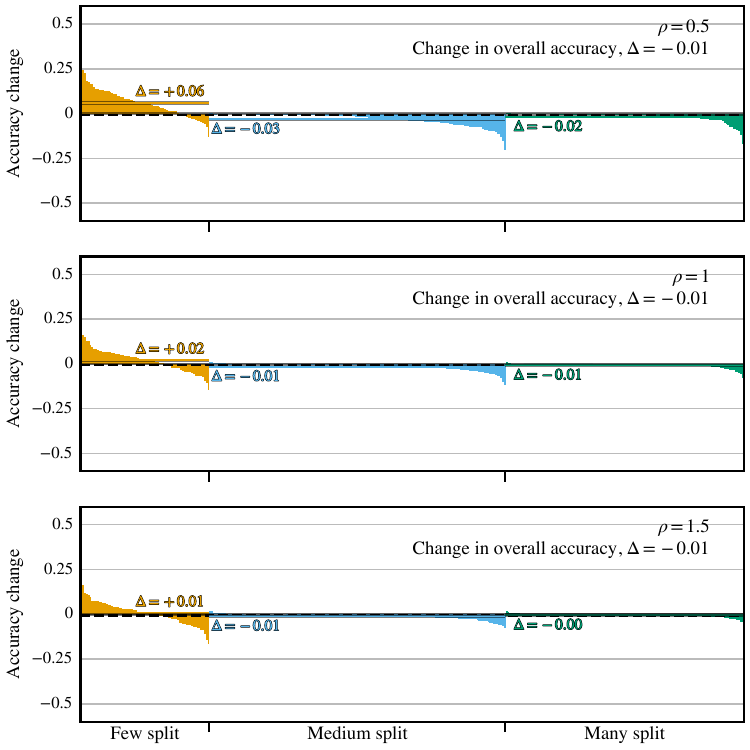}
\caption{Change in per-class test accuracy on Places‑LT after AlphaNet training with cRT baseline. Each bar shows the change the change in accuracy for one class. The solid lines in each split show the average per-class change for the split, and the dotted line shows the overall average per-class change.}\label{fig:rhos_cls_deltas_placeslt_crt}
}
\end{figure}

\begin{figure}
\hypertarget{fig:rhos_cls_deltas_placeslt_lws}{%
\centering
\includegraphics{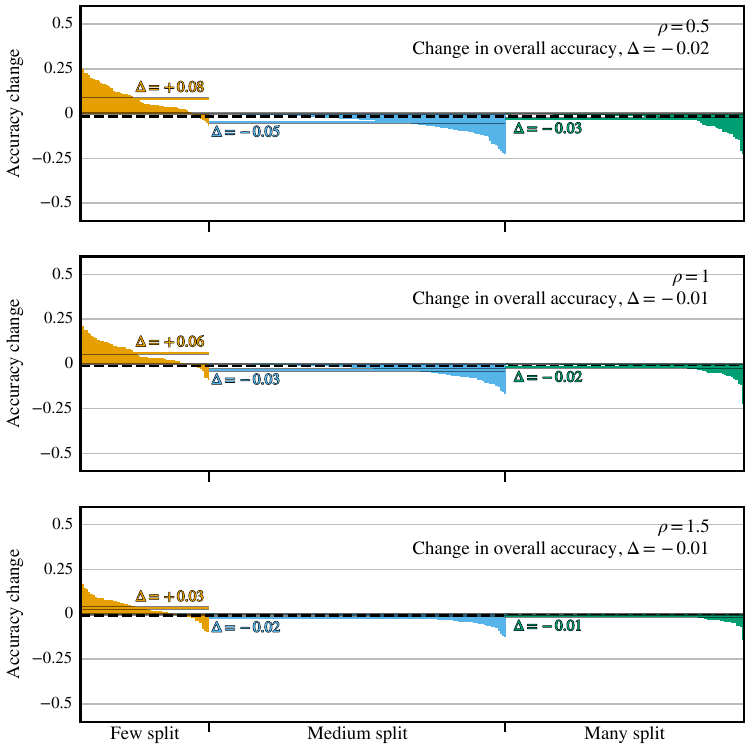}
\caption{Change in per-class test accuracy on Places‑LT after AlphaNet training with LWS baseline. Each bar shows the change the change in accuracy for one class. The solid lines in each split show the average per-class change for the split, and the dotted line shows the overall average per-class change.}\label{fig:rhos_cls_deltas_placeslt_lws}
}
\end{figure}

\begin{figure}
\hypertarget{fig:rhos_cls_deltas_cifarlt_ride}{%
\centering
\includegraphics{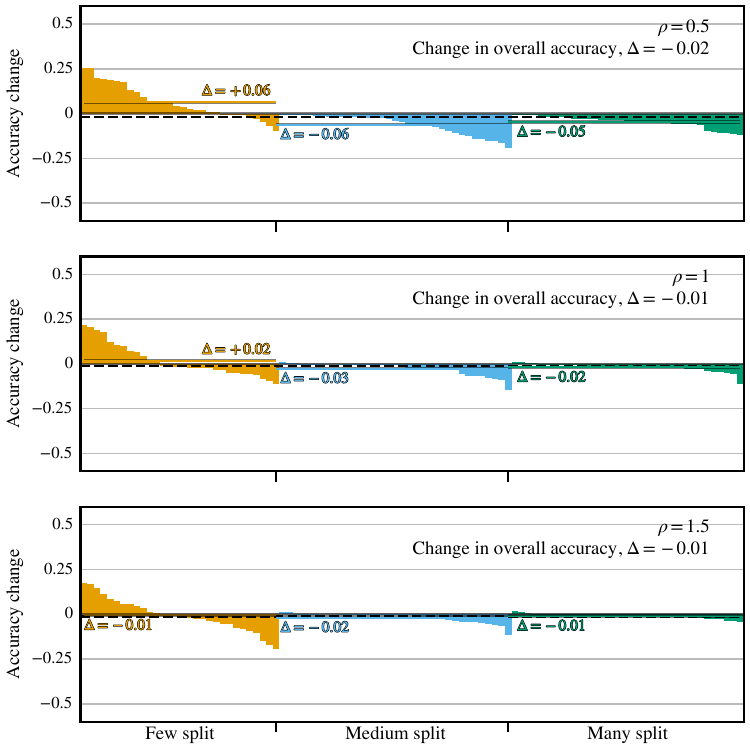}
\caption{Change in per-class test accuracy on CIFAR‑100‑LT after AlphaNet training with RIDE baseline. Each bar shows the change the change in accuracy for one class. The solid lines in each split show the average per-class change for the split, and the dotted line shows the overall average per-class change.}\label{fig:rhos_cls_deltas_cifarlt_ride}
}
\end{figure}

\begin{figure}
\hypertarget{fig:rhos_cls_deltas_cifarlt_ltr}{%
\centering
\includegraphics{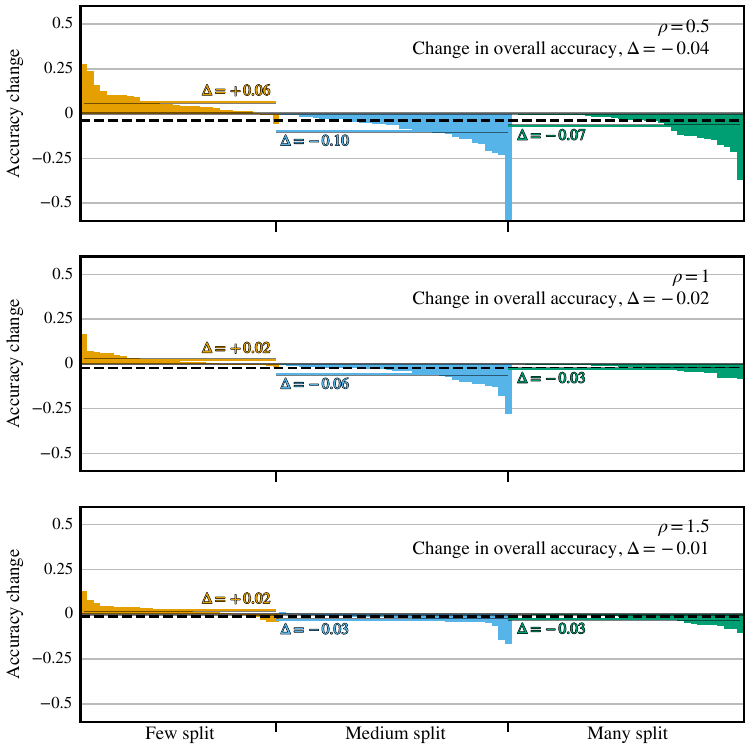}
\caption{Change in per-class test accuracy on CIFAR‑100‑LT after AlphaNet training with LTR baseline. Each bar shows the change the change in accuracy for one class. The solid lines in each split show the average per-class change for the split, and the dotted line shows the overall average per-class change.}\label{fig:rhos_cls_deltas_cifarlt_ltr}
}
\end{figure}

In this section, we analyze the change in accuracy for individual classes after applying AlphaNet (following the training process described in Section~\ref{sec:impl}). First, we plotted the sorted accuracy changes, grouped by split. These are shown in the following figures:

\begin{itemize}
\tightlist
\item
  ImageNet‑LT:

  \begin{itemize}
  \tightlist
  \item
    cRT baseline: Figure~\ref{fig:rhos_cls_deltas_imagenetlt_crt}.
  \item
    LWS baseline: Figure~\ref{fig:rhos_cls_deltas_imagenetlt_lws}.
  \item
    RIDE baseline: Figure~\ref{fig:rhos_cls_deltas_imagenetlt_ride}.
  \end{itemize}
\item
  Places‑LT:

  \begin{itemize}
  \tightlist
  \item
    cRT baseline: Figure~\ref{fig:rhos_cls_deltas_placeslt_crt}.
  \item
    LWS baseline: Figure~\ref{fig:rhos_cls_deltas_placeslt_lws}.
  \end{itemize}
\item
  CIFAR‑100‑LT:

  \begin{itemize}
  \tightlist
  \item
    RIDE baseline: Figure~\ref{fig:rhos_cls_deltas_cifarlt_ride}.
  \item
    LTR baseline: Figure~\ref{fig:rhos_cls_deltas_cifarlt_ltr}.
  \end{itemize}
\end{itemize}

\begin{figure}
\subfloat[cRT baseline]{\includegraphics{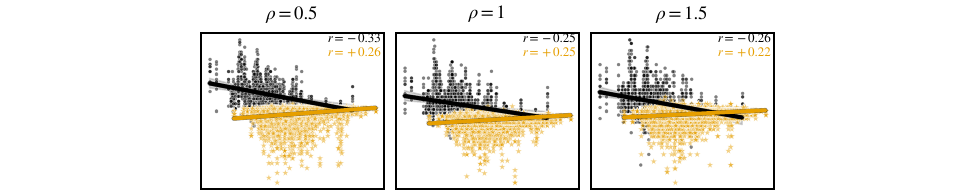}\label{fig:rhos_cls_delta_vs_nndist_imagenetlt:crt}}
\quad
\subfloat[LWS baseline]{\includegraphics{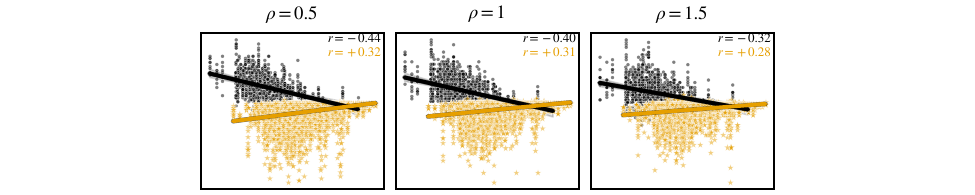}\label{fig:rhos_cls_delta_vs_nndist_imagenetlt:lws}}
\quad
\subfloat[RIDE baseline]{\includegraphics{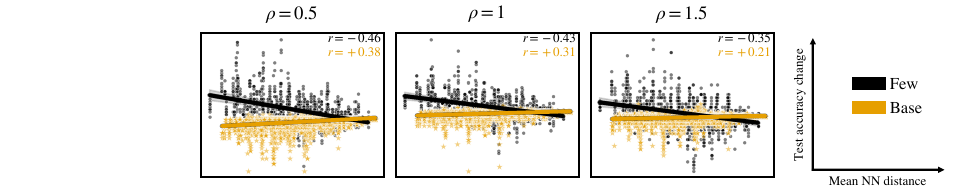}\label{fig:rhos_cls_delta_vs_nndist_imagenetlt:ride}}
\caption{Change in per-class test accuracy on ImageNet‑LT, versus mean distance to 5 nearest neighbors based on Euclidean distance. The neighbors are from `base' split for the `few' split classes, and vice-versa for the `base' split classes. The lines are regression fits, and the `\(r\)'s are Pearson correlations.}
\label{fig:rhos_cls_delta_vs_nndist_imagenetlt}
\end{figure}

\begin{figure}
\subfloat[cRT baseline]{\includegraphics{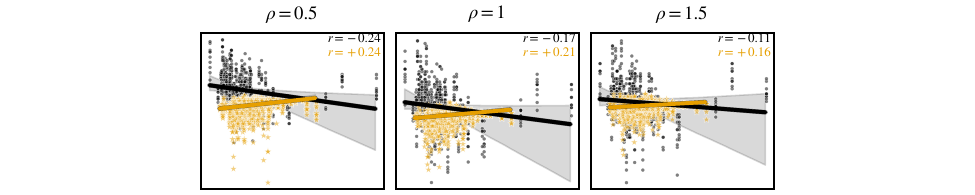}\label{fig:rhos_cls_delta_vs_nndist_placeslt:crt}}
\quad
\subfloat[LWS baseline]{\includegraphics{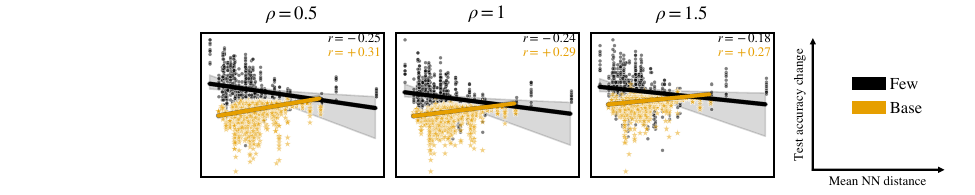}\label{fig:rhos_cls_delta_vs_nndist_placeslt:lws}}
\caption{Change in per-class test accuracy on Places‑LT, versus mean distance to 5 nearest neighbors based on Euclidean distance. The neighbors are from `base' split for the `few' split classes, and vice-versa for the `base' split classes. The lines are regression fits, and the `\(r\)'s are Pearson correlations.}
\label{fig:rhos_cls_delta_vs_nndist_placeslt}
\end{figure}

\begin{figure}
\subfloat[RIDE baseline]{\includegraphics{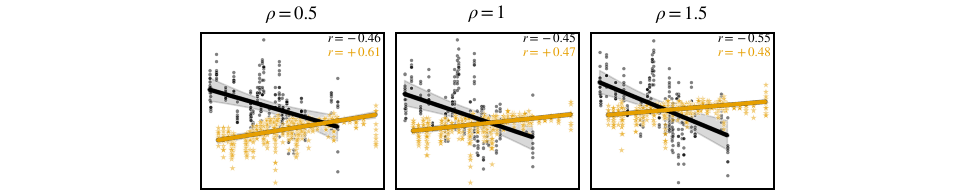}\label{fig:rhos_cls_delta_vs_nndist_cifarlt:ride}}
\quad
\subfloat[LTR baseline]{\includegraphics{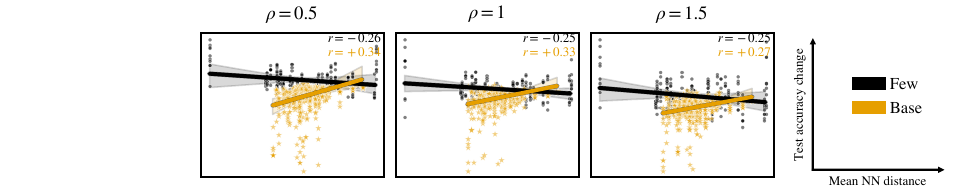}\label{fig:rhos_cls_delta_vs_nndist_cifarlt:ltr}}
\caption{Change in per-class test accuracy on CIFAR‑100‑LT, versus mean distance to 5 nearest neighbors based on Euclidean distance. The neighbors are from `base' split for the `few' split classes, and vice-versa for the `base' split classes. The lines are regression fits, and the `\(r\)'s are Pearson correlations.}
\label{fig:rhos_cls_delta_vs_nndist_cifarlt}
\end{figure}

We also plotted accuracy change for classes against the average distance to their 5 nearest neighbors by Euclidean distance. For `few' split classes, we selected neighbors from the `base' split, and for the `base' split classes, we selected neighbors from the `few' split. Recall that classifiers for the `few' split were updated using classifiers from these neighbors. Results are shown in the following figures:

\begin{itemize}
\tightlist
\item
  ImageNet‑LT:

  \begin{itemize}
  \tightlist
  \item
    cRT baseline: Figure~\ref{fig:rhos_cls_delta_vs_nndist_imagenetlt:crt}.
  \item
    LWS baseline: Figure~\ref{fig:rhos_cls_delta_vs_nndist_imagenetlt:lws}.
  \item
    RIDE baseline: Figure~\ref{fig:rhos_cls_delta_vs_nndist_imagenetlt:ride}.
  \end{itemize}
\item
  Places‑LT:

  \begin{itemize}
  \tightlist
  \item
    cRT baseline: Figure~\ref{fig:rhos_cls_delta_vs_nndist_placeslt:crt}.
  \item
    LWS baseline: Figure~\ref{fig:rhos_cls_delta_vs_nndist_placeslt:lws}.
  \end{itemize}
\item
  CIFAR‑100‑LT:

  \begin{itemize}
  \tightlist
  \item
    RIDE baseline: Figure~\ref{fig:rhos_cls_delta_vs_nndist_cifarlt:ride}.
  \item
    LTR baseline: Figure~\ref{fig:rhos_cls_delta_vs_nndist_cifarlt:ltr}.
  \end{itemize}
\end{itemize}

\clearpage

\clearpage

\clearpage

\clearpage

\clearpage

\clearpage

\clearpage

\clearpage

\clearpage

\clearpage

\clearpage